\documentclass[10pt,twocolumn,letterpaper]{article}

\usepackage[pagenumbers]{cvpr} 

\usepackage{graphicx}
\usepackage{amsmath}
\usepackage{amssymb}
\usepackage{diagbox}
\usepackage{bbding}
\usepackage{booktabs}
\usepackage{makecell}
\usepackage{bm}
\usepackage{url}
\usepackage{pifont}
\usepackage{multirow}
\usepackage{tabularx,arydshln}

\usepackage[pagebackref,breaklinks,colorlinks]{hyperref}

\usepackage[capitalize]{cleveref}
\crefname{section}{Sec.}{Secs.}
\Crefname{section}{Section}{Sections}
\Crefname{table}{Table}{Tables}
\crefname{table}{Tab.}{Tabs.}

\def\cvprPaperID{2814} 
\def\confName{CVPR}
\def\confYear{2022}
\begin{document}

\title{Group Contextualization for Video Recognition}

\author{Yanbin Hao\\
University of Science and Technology of China\\
Hefei, China\\
{\tt\small haoyanbin@hotmail.com}
\and
Hao Zhang\thanks{Hao Zhang is the corresponding author.}\\
Singapore Management University\\
Singapore\\\
{\tt\small hzhang@smu.edu.sg}
\and
Chong-Wah Ngo\\
Singapore Management University\\
Singapore\\
{\tt\small cwngo@smu.edu.sg}
\and
Xiangnan He\\
University of Science and Technology of China\\
Hefei, China\\
{\tt\small xiangnanhe@gmail.com}
}
\maketitle

\begin{abstract}
\vspace{-0.2cm}
Learning discriminative representation from the complex spatio-temporal dynamic space is essential for video recognition. On top of those stylized spatio-temporal computational units, further refining the learnt feature with axial contexts is demonstrated to be promising in achieving this goal. However, previous works generally focus on utilizing a single kind of contexts to calibrate entire feature channels and could hardly apply to deal with diverse video activities. The problem can be tackled by using pair-wise spatio-temporal attentions to recompute feature response with cross-axis contexts at the expense of heavy computations. In this paper, we propose an efficient feature refinement method that decomposes the feature channels into several groups and separately refines them with different axial contexts in parallel. We refer this lightweight feature calibration as group contextualization (GC). Specifically, we design a family of efficient element-wise calibrators, i.e., ECal-G/S/T/L, where their axial contexts are information dynamics aggregated from other axes either globally or locally, to contextualize feature channel groups. The GC module can be densely plugged into each residual layer of the off-the-shelf video networks. With little computational overhead, consistent improvement is observed when plugging in GC on different networks. By utilizing calibrators to embed feature with four different kinds of contexts in parallel, the learnt representation is expected to be more resilient to diverse types of activities. On videos with rich temporal variations, empirically GC can boost the performance of 2D-CNN (e.g., TSN and TSM) to a level comparable to the state-of-the-art video networks. Code is available at \url{https://github.com/haoyanbin918/Group-Contextualization}.
\end{abstract}

\vspace{-0.2cm}
\vspace{-0.2cm}
\section{Introduction}

\begin{figure}[t]
\centering
\includegraphics[width=0.46\textwidth]{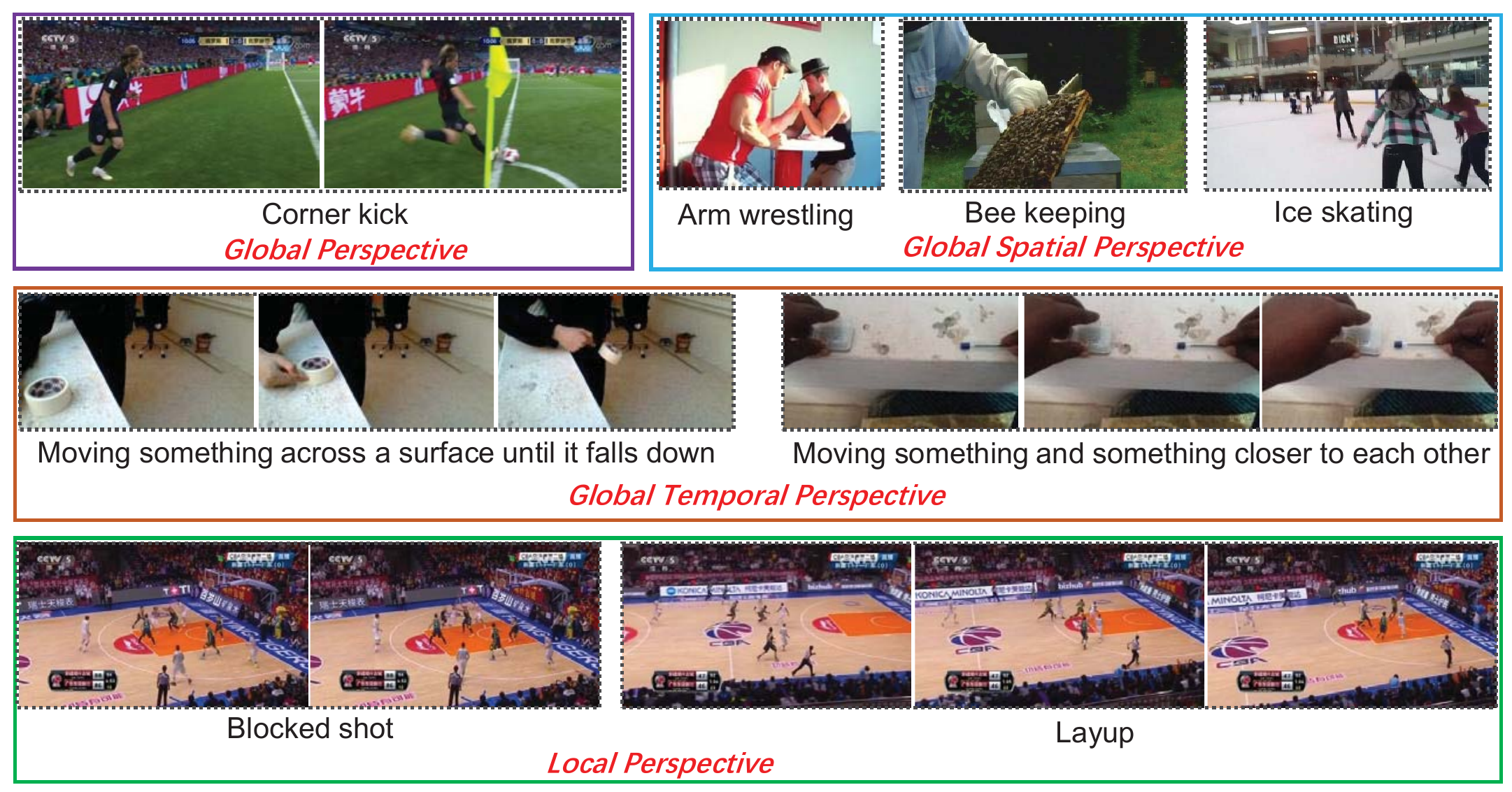}
\vspace{-0.2cm}
\caption{Perspective/axial preference of different video activities. The scene change caused by quick camera movement yearns for global context for recognizing the soccer highlight ``\textit{Corner kick}''. ``\textit{Arm wrestling}'',  ``\textit{Bee keeping}'' and ``\textit{Ice skating}'' can be easily recognized even by a single keyframe. Whereas, the Something-Something activity examples (middle) rely much on temporal relations. The group activities, i.e., ``\textit{Blocked shot}'' and ``\textit{Layup}'', require a model to localize sub-activities.}
\label{demos}
\vspace{-0.5cm}
\end{figure}

The 3D spatio-temporal nature of video signals allows video content to be flexibly analyzed from different perspectives or axes. Specifically, the signals can be transformed along various dimensions to capture the activities underlying a video. For example, in Figure \ref{demos}, the soccer highlight ``\textit{Corner kick}'' may require projection of the 3D video signal into a 1D vector to globally summarize the quick camera movement causing scene change. The less temporal activity categories ``\textit{Arm wrestling}'', ``\textit{Bee keeping}'' and ``\textit{Ice skating}'' occurring in near static scenes prefer a 2D image representation to capture spatial perspective for classification. By contrast, the video events ``\textit{Moving something across a surface until it falls down}'' and ``\textit{Moving something and something closer to each other}'' require modeling of temporal relations over time axis. When it is about classifying sub-activities such as ``\textit{Blocked shot}'' and ``\textit{Layup}'' in a lengthy basketball video, the localized 3D spatio-temporal analysis is preferred. These observations show the necessity of adjusting the features ($C$ channels) of a 3-dimensional $T\times H\times W$ video tensor with a perspective aligned with video activities.

Feature contextualization \cite{wang2018non,hao2020compact,xie2018rethinking,hu2018squeeze,li2020tea,liu2020tam} is a technique that makes full use of axial contexts (e.g., spatial, temporal) to calibrate plain video features obtained from convolutional filters of CNN models (e.g., C3D~\cite{tran2015learning}, I3D~\cite{carreira2017quo}, P3D~\cite{qiu2017learning}). Generally, axial contexts are referred to as information aggregated from other axes towards the features. For example, the global spatial context within the whole time length can be obtained by squeezing the tensor along time axis \cite{xie2018rethinking}, while in contrast the global temporal context is acquired through shrinking along space axes \cite{li2020tea,liu2020tam}. However, due to the large diversity of video activities, it is obvious that a single context cannot fit all activity cases. Given the activity ``\textit{Moving something and something closer to each other}'' in Figure \ref{demos} as an example, globally aggregating contents along the time axis will harm the time order information, underemphasizing the subtle movement between objects. Also, the global aggregation will mess up sub-activities in the basketball highlights, diminishing the characteristics localized to sub-activities such as ``dunk'' and ``foul''. In these cases, the existing works \cite{xie2018rethinking,hu2018squeeze,li2020tea,liu2020tam}, which focus on calibrating image/video features with only a specific global axial context, may under-perform due to the lack of versatility in representing various activities. On the other hand, projecting multiple axial contexts to a feature will increase the computation cost. Some works \cite{wang2018non,hao2020compact} try to pairwisely attend each feature point of the 3D video tensor from local to global receptive field to adaptively decide the perspectives depending on context. Nevertheless, these works suffer from the heavy computation burden. The resulting network is not lightweight, and cannot densely plug into the existing network backbone. The most current work temporal difference network (TDN) \cite{wang2021tdn} shows strong performance on activities that require both short-term and long-term dependencies through pairwisely computing the temporal differences with short and long intervals of time.

\begin{figure}[t]
\centering
\includegraphics[width=0.48\textwidth]{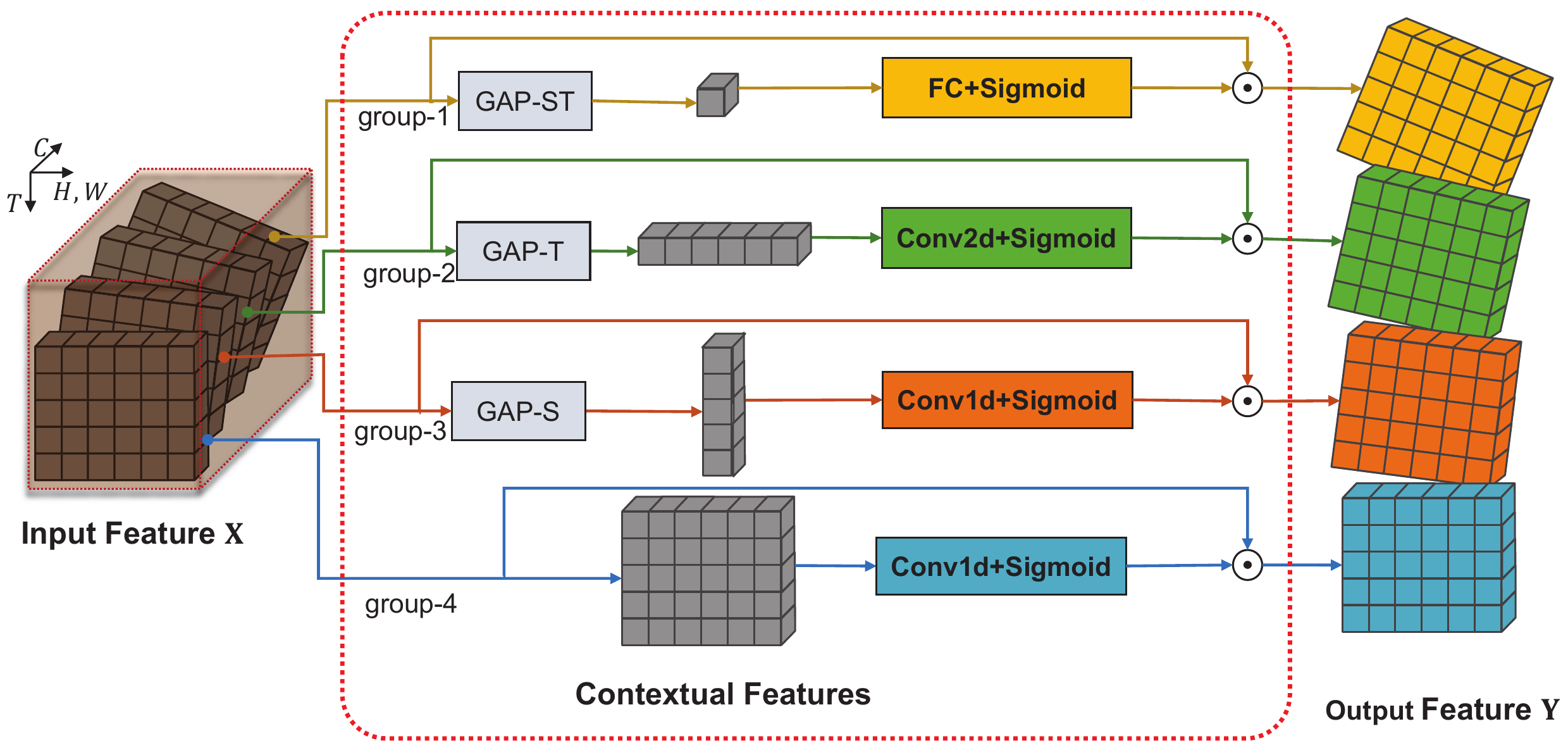}
\vspace{-0.5cm}
\caption{The workflow of the proposed group contextualization. ``GAP'' denotes global average pooling.}
\label{pipeline}
\vspace{-0.5cm}
\end{figure}

This paper addresses the limitation of feature contextualization for video recognition.~Specifically, a novel feature contextualization paradigm, i.e, group contextualization (GC), is presented to derive feature representation that is generic to different activities and with lightweight computation.~The GC module decomposes the channels into several paralleled groups and applies different feature contextualization operations on them respectively. As such, the calibrated feature is versatile for it integrating feature dynamics aggregated from different perspectives and potentially can recognize a wide variety of activities. The computational overload is kept to a minimum level by applying group convolution \cite{xie2017aggregated,luo2019grouped,tran2019video}. As each output channel only relates to the input channels within a group, only $\frac{1}{g}C^2$ ($g$ is the number of divided groups) channel interactions is required, instead of $C^2$ of a standard convolution. Capitalizing on efficiency in computation, group convolution also allows us to analyze how a network exploits axial contexts in different layers for different activities.

The workflow of GC module is illustrated in Figure \ref{pipeline}. Particularly, the input CNN feature $\textbf{X}\in \mathbb{R}^{T\times H\times W\times C}$ is split into four groups (group-1/2/3/4 with the size of  $T\times H\times W\times \frac{1}{4}C$) along the channel dimension. To achieve separate contextualization, we accordingly design four element-wise calibrators\footnote{In this paper, feature adjustment, refinement and calibration, as well as their noun and verb forms, are used interchangeably.} (ECals) to calibrate the four channel groups globally along space-time (ECal-G), globally along space (ECal-S), globally along time (ECal-T), and locally (ECal-L) in a small neighborhood in parallel. In this architecture, all ECals share the similar cascaded structure of ``GAP/None+FC/Conv+Sigmoid'' for efficiency, and achieve feature calibration with element-wise multiplication. Here, the global feature calibrations (ECal-G/S/T) perform feature pooling along a specific axis and contextualization on the different axis with the customized operators. In fact, the feature aggregation compresses the global information in an axis onto the other so that it enlarges the receptive field to the entire axial range. For example, ECal-T squeezes the input $T\times H\times W\times \frac{1}{4}C$ feature along the space axes, resulting in a $T\times 1\times 1\times \frac{1}{4}C$ contextual feature. In this case, when conducting temporal convolution on the resulted context feature, the global spatial content contributes the attention weight computation of each timestamp, which can benefit the recognition of video activities requiring long-range temporal relation (e.g., the video clips in Figure \ref{demos}(middle)). In contrast, if we directly convolve the given feature map without any pooling operation, the global view narrows to a local neighbourhood (ECal-L). This perspective is particularly useful to localize sub-activities in lengthy video, e.g., ``\textit{Blocked shot}'', ``\textit{Layup}'' in the basketball video shown in Figure \ref{demos}. Finally, by separately performing feature refinement with ECal-G/S/T/L on those decomposed channel groups, we can reweight the input feature with multiple axial perspectives, resulting in a more discriminative representation $\textbf{Y}$.

We summarize our contributions as below:
\vspace{-0.1cm}
\begin{itemize}
\item \textbf{Group contextualization}.~We propose a new regime named group contextualization (GC) for video feature refinement. GC encompasses a set of element-wise calibrators (ECal-G/S/T/L) to explicitly model multi-axial contexts and separately refine video feature groups in parallel. 

\vspace{-0.1cm}
\item \textbf{Computation-efficient}.~All ECal variants are designed in an efficient manner, and the channel decomposition as in group convolution moderates the extra computational cost incurred in feature calibration. For example, when averagely splitting the channels into four groups, GC only introduces  5.3\%/1.3\% extra parameters/FLOPs to the original TSN backbone.

\vspace{-0.1cm}
\item \textbf{Significant performance gain}.~We verify that our GC module not only significantly improves the video recognition performance for several feedforward video networks (i.e., TSN, TSM and GST), but also can work together with the temporal difference network (TDN) leading to a notable performance gain.
\end{itemize}

\section{Related Work}
Since our work is mainly relevant to feature contextualization and group convolution techniques, we will separately review  the related works from these two aspects.

{\bf Feature contextualization}. Feature contextualization has been successfully demonstrated to be effective in image and video processing tasks, such as image retrieval/classification/segmentation~\cite{zhu2019r2gan,guo2020visual,han2021fine,zhu2020cookgan,wang2017residual,zhang2018fine,hu2018squeeze,misra2021rotate,fu2019dual}, video/action recognition/classification \cite{wang2018non,xie2018rethinking,hao2020compact,li2020tea,liu2020tam,guo2021ssan,tan2021selective}. Contextualization operation can enlarge the local receptive filed of a spatio-temporal filter to a global view with the support of perspective contexts. For example, the non-local neural network~\cite{wang2018non} recomputes the output of a local filter as a weighted sum of features of all points in the whole spatio-temporal video space.  Since this operation needs pairwise comparison, its power is limited by the heavy computation burden. SSAN~\cite{guo2021ssan} factorizes the 3D pairwise attention into three separable spatial, temporal and channel attentions for efficiency. Another similar work is CBA-QSA~\cite{hao2020compact} which omits the pairwise comparison and instead introduces a learnable query to guide the attention weight computing. To achieve more efficient feature contextualization, some approaches have studied modeling axial contexts through squeezing along specific axes. For example, SE-Net firstly proposes the squeeze-and-excitation mechanism to work as a self-gating operator to elementwisely refine image features with global context. The gather-excite network (GE-Net) \cite{hu2018gather} generalizes SE-Net by investigating various levels of spatial context granularity. S3D-G~\cite{xie2018rethinking} brings the feature refinement idea of SE-Net \cite{hu2018squeeze} to calibrate the features of S3D with the global axial context. TEA~\cite{li2020tea} introduces a motion excitation module to calculate pixel-wise movement of subsequent frames and a multiple temporal aggregation module to enlarge the temporal receptive field with the aggregated temporal axial context. TANet~\cite{liu2020tam} also performs average pooling to collapse spatial information towards time axis but additionally considers long-range temporal modeling by having a feedforward neural network as a separate branch to refine the features. Compared to prior works, GC not only takes the global/temporal axial context into account for long-range temporal modeling but also considers the underlying video activities for feature contextualization.

{\bf Group convolution}. The group convolution \cite{krizhevsky2012imagenet,xie2017aggregated,zhang2018shufflenet,hara2018can,tran2019video,luo2019grouped} divides the feature maps into small groups and uses multiple kernels to separately compute their channel outputs. This leads to not only much lower computation loads but also wider networks helping to learn a varied set of low level and high level features. In video processing area, the work \cite{hara2018can}, which directly replaces the spatial 2D convolutional kernels of 2D-CNNs such as ResNext~\cite{xie2017aggregated} and DenseNet~\cite{huang2017densely} with 3D counterparts, explores the potential of 3D group convolutions for video recognition. The channel-separated convolutional network (CSN) \cite{tran2019video} studies various settings of 3D group convolution as well as its extreme version depthwise convolution on C3D~\cite{tran2015learning} for efficient video classification. More recently, the grouped spatial-temporal network (GST) \cite{luo2019grouped} proposes to decompose the feature channels into two asymmetric groups and uses 2D and 3D convolutions to separately learn the spatial and temporal information. The gate-shift module (GSM) \cite{sudhakaran2020gate} extends temporal shift module used in \cite{lin2019tsm,zhang2021token} with learnable shift parameters and uses the channel decomposition to further reduce parameters. The above group convolution works focus on designing generic models, while our proposed group contextualization is to recalibrate the plain video feature with multiple axial contexts for enhancing the off-the-shelf neural network models for video recognition.

\section{Group Contextualization}

The group contextualization module is constructed as a plug-and-play module,  which can be used to calibrate any given 4D  $T\times H\times W\times C$ video tensor. In this section, we first elaborate the details of GC module, as well as four designed element-wise calibrators, in a general manner. Then, we integrate it into four representative video CNN models for enhancing their capacity of representation learning and give analysis to model complexity. Finally we examine the impact of varying channel positions in the backbones.

\begin{figure}[t]
\centering
\includegraphics[width=0.45\textwidth]{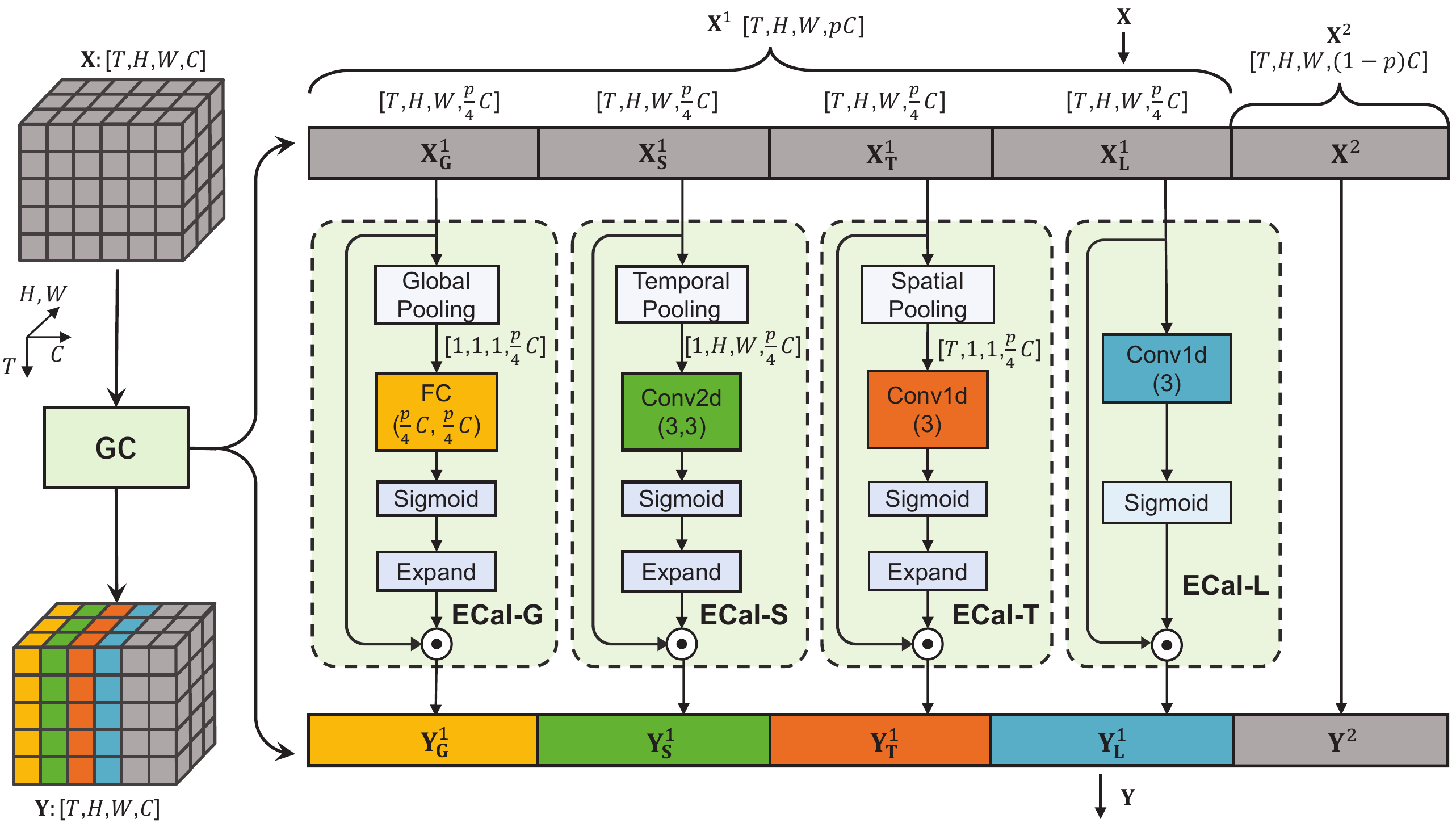}
\vspace{-0.2cm}
\caption{The schema of the GC module. ``$\odot $'' denotes the Hadamard product. }
\label{model_struc}
\vspace{-0.4cm}
\end{figure}

GC aims to calibrate a portion of channels of a video feature using a specific axial context at a time. The schema of GC module is illustrated in Figure \ref{model_struc}. Suppose that a 4D feature tensor is $\textbf{X}\in \mathbb{R}^{T\times H\times W\times C}$ yielded by a convolutional operator or counterpart, where $T,H,W,C$ denote time-length, space-height, space-width and channel-size, respectively. Firstly, GC splits the feature tensor into two groups with a partition ratio $p$ along the channel dimension, resulting in $\textbf{X}^{1}\in \mathbb{R}^{T\times H\times W\times pC}$ and $\textbf{X}^{2}\in \mathbb{R}^{T\times H\times W\times (1-p)C}$. Then, four feature calibrators are customized to focus on four different axial perspectives and separately refine the four feature channel subgroups of $\textbf{X}^{1}$, i.e., $\textbf{X}^{1}_{G/S/T/L}\in \mathbb{R}^{T\times H\times W\times \frac{p}{4}C}$ in parallel, resulting in four corresponding outputs $\textbf{Y}^{1}_{G/S/T/L}\in \mathbb{R}^{T\times H\times W\times \frac{p}{4}C}$. Finally, the calibrated feature parts and the non-calibrated feature part are concatenated along channel dimension, and the output of GC is
\begin{equation}
    \textbf{Y} = \text{Concat}(\textbf{Y}^{1}_{G},\textbf{Y}^{1}_{S},\textbf{Y}^{1}_{T},\textbf{Y}^{1}_{L},\textbf{Y}^{2}) ,  \\ 
\end{equation}
where $\textbf{Y}\in \mathbb{R}^{T\times H\times W\times C}$. Next, we present the designs of four element-wise calibrators in detail.

\subsection{Element-wise Calibrators}

{\bf Global-wise Calibrator (ECal-G)}. The ECal-G block instantiates the global axial context through globally pooling the 4D $\textbf{X}^{1}_{G}$ across time and space, yielding a contextual vector with the size of $\frac{p}{4}C$. Then, to make use of the aggregated contextual information, a fully-connected (FC) layer is to compute the interactions among channels of the vector. Finally, a Sigmoid function is employed to calculate the channel-wise gating weights and an Expand operation  further inflates the weight vector to the same size of $\textbf{X}^{1}_{G}$ by element copying. Formally, the calculation flow can be as follows
\begin{small}
\begin{equation}
    \textbf{Y}^{1}_{G}=\text{Expand} (\text{Sigmoid}(\text{FC} (\frac{1}{T\times H\times W}\sum_{t,h,w} \textbf{X}_{G}^{1}[t,h,w] ) ) )\odot \textbf{X}^{1}_{G}.
\end{equation}
\end{small}

{\bf Spatial-wise Calibrator (ECal-S)}. ECal-S block shrinks the input tensor along the temporal axis using an average pooling operation, resulting in a $1\times H\times W\times C$ contextual feature. A 2D convolution with $3\times 3$ kernel is then adopted to compute the impact to a local spatial neighbor. Similarly, the Sigmoid and Expand operations are for the element-wise weighting tensor generation. Finally, we have
\begin{small}
\begin{equation}
    \textbf{Y}^{1}_{S}=\text{Expand} (\text{Sigmoid}(\text{Conv2d} (\frac{1}{T}\sum_{t}\textbf{X}^{1}_{S}[t,:,:] ) ) )\odot \textbf{X}^{1}_{S}.
\end{equation}
\end{small}

{\bf Temporal-wise Calibrator (ECal-T)}. In contrast to ECal-S, ECal-T pools the input along the spatial axes, aggregating the global spatial content into $T\times 1\times 1\times C$ statistics. Feature contextualization is achieved by using a temporal 1D convolution, which can mix the global spatial information within a local temporal receptive field. Further passing to the Sigmoid and Expand functions, the refined output tensor $\textbf{Y}^{1}_{G}$ is computed as 
\begin{small}
\begin{equation}
    \textbf{Y}^{1}_{T}=\text{Expand} (\text{Sigmoid} (\text{Conv1d} (\frac{1}{H\times W}\sum_{h,w}\textbf{X}^{1}_{T}[:,h,w] ) ) )\odot \textbf{X}^{1}_{T}.
\end{equation}
\end{small}

{\bf Local-wise Calibrator (ECal-L)}. Since ECal-L focuses on capturing local contexts within a neighboring field, we directly utilize a convolutional unit to achieve the local interaction computation. In the implementation, a 1D temporal convolution with $3\times 1\times 1$ kernel is adopted. This is because that a simple 1D convolution requires much lighter computational load than a 3D convolution, and temporal modeling is more critical in video feature learning. Without the use of global average pooling operation, the size of input tensor in ECal-L is kept during the weight calculation. Hereby, we have
\begin{small}
\begin{equation}
    \textbf{Y}^{1}_{L}=\text{Sigmoid} (\text{Conv1d} (\textbf{X}^{1}_{L} ) )\odot \textbf{X}^{1}_{L}.
\end{equation}
\end{small}

The above four ECals work individually and follow the self-gating regime for feature calibration. They compute the element-wise gating weights by contextualizing a specific axial perspective of interest. The element-wise gating weights could be global-wise (ECal-G),spatial-wise (ECal-S), temporal-wise (ECal-T), and local-wise (ECal-L). As a result, the proposed GC module can achieve multiple perspectives of feature contextualization in parallel for a single input.

\subsection{Network Architecture and Model Complexity}

\begin{figure}
\centering
\includegraphics[width=60mm]{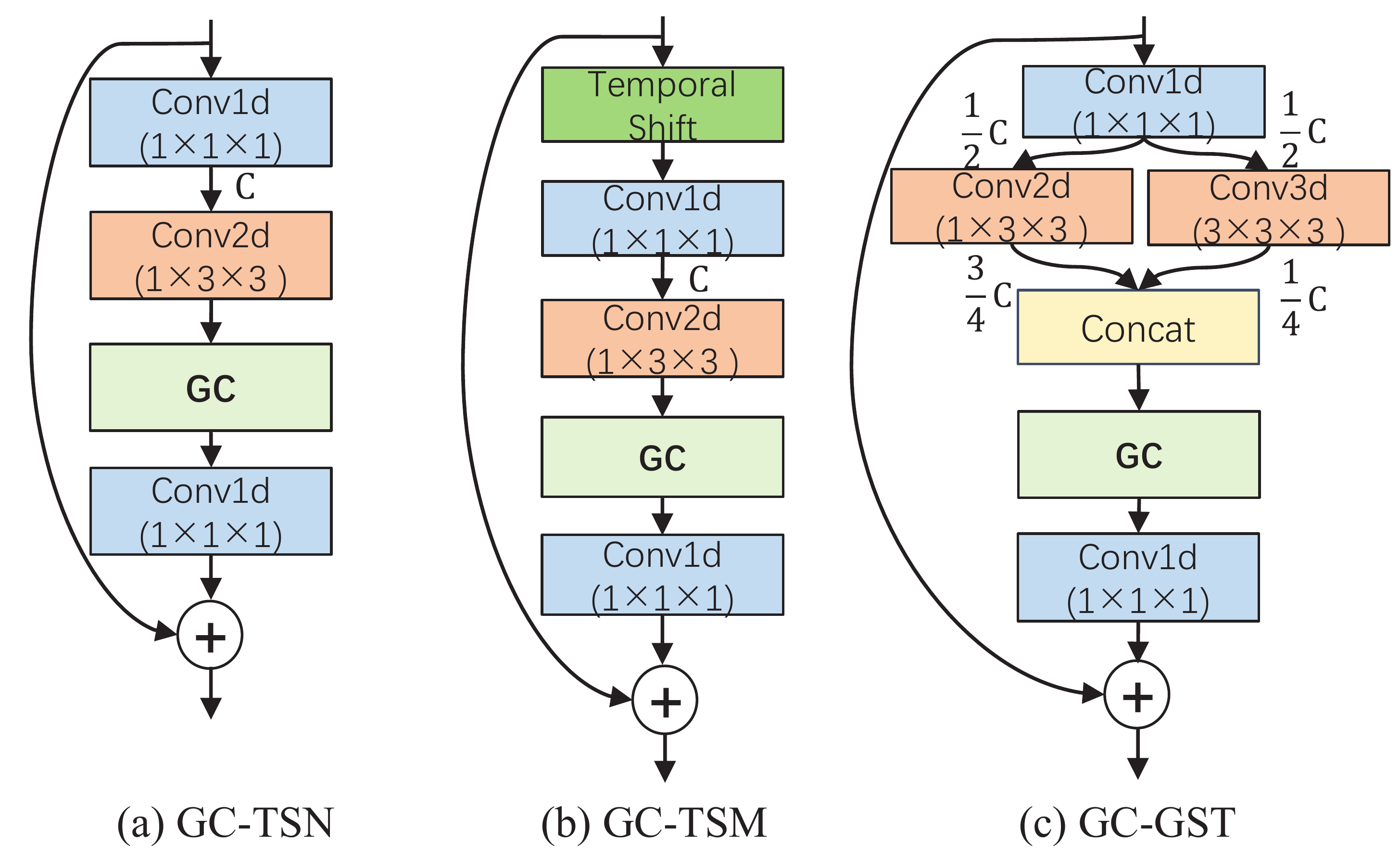}
\vspace{-0.2cm}
\caption{Integration of GC module into TSN, TSM and GST networks.}
\label{tsn-tsm-csa}
\vspace{-0.6cm}
\end{figure}

We integrate the GC module into three basic video networks, i.e., TSN~\cite{wang2016temporal} (a  standard 2D spatial model), TSM~\cite{lin2019tsm} (a 2D temporal shift model), and GST~\cite{luo2019grouped} (a 3D group convolution model), and a more advanced network, i.e., TDN~\cite{wang2021tdn} (modeling both short-term and long-term dependencies by temporal differences), referred to as GC-TSN, GC-TSM, GC-GST and GC-TDN. Figure \ref{tsn-tsm-csa} shows the integrated blocks in GC-TSN/TSM/GST. The GC module is inserted after the 2D/3D convolution layer. Particularly, since TDN uses different stages to separately model the short-term (the first two stages) and long-term temporal information (the latter three stages), we thus do not visualize its integrated block. The implementation of GC-TDN follows GC-TSN.

\begin{table}[htbp]
		\centering
		\scriptsize
		\begin{tabular}{l|cccc}
			\hline
			\multirow{2}{*}{Block} &\multirow{2}{*}{Params} &&\multicolumn{2}{c}{Percentage}  \\
			\cline{4-5}
			 &   && $p=\frac{1}{2}$    &$p=1$  \\
            \midrule[1pt]
            Residual block (TSN) &$17\times C^{2}$  &&\multicolumn{2}{c}{100.0\%}   \\
            \hline
            ECal-G &$\frac{1}{16}\times p^2 \times C^{2}$ &&0.09\%  &0.37\%  \\
            ECal-S &$\frac{9}{16}\times p^2 \times C^{2}$ &&0.83\%  &3.31\%   \\
            ECal-T &$\frac{3}{16}\times p^2 \times C^{2}$ &&0.28\%  &1.10\%   \\
            ECal-L &$\frac{3}{16}\times p^2 \times C^{2}$ &&0.28\%  &1.10\%   \\
            Total &$p^2\times C^{2}$ &&1.47\%  &5.88\% \\
            \hline
		\end{tabular}
		\vspace{-0.2cm}
		\caption{Comparison of parameters for different calibrator blocks. For simplicity, bias terms and batchnorm terms are omitted in parameter counting.}
		\label{tab:compcost}
	\vspace{-0.4cm}
\end{table}

The partition ratio $p$ controls the portion of channels to be calibrated and hence governs the complexity of GC module. Table \ref{tab:compcost} lists the number of parameters of each ECal as well as their sums. To make a clearer comparison, we also show the number of parameters of the original residual block of TSN. It is worth noting that TSN and TSM have the same model complexity as the temporal shift operation in TSM is computationally free. Specifically, the parameters introduced by GC module is as low as 1.47\% of the original 2D Residual block when $p=\frac{1}{2}$.


\subsection{Does Channel Position Make Any Difference?}
\label{channelposition}

\begin{figure}[tbp]
\centering
\includegraphics[width=0.4\textwidth]{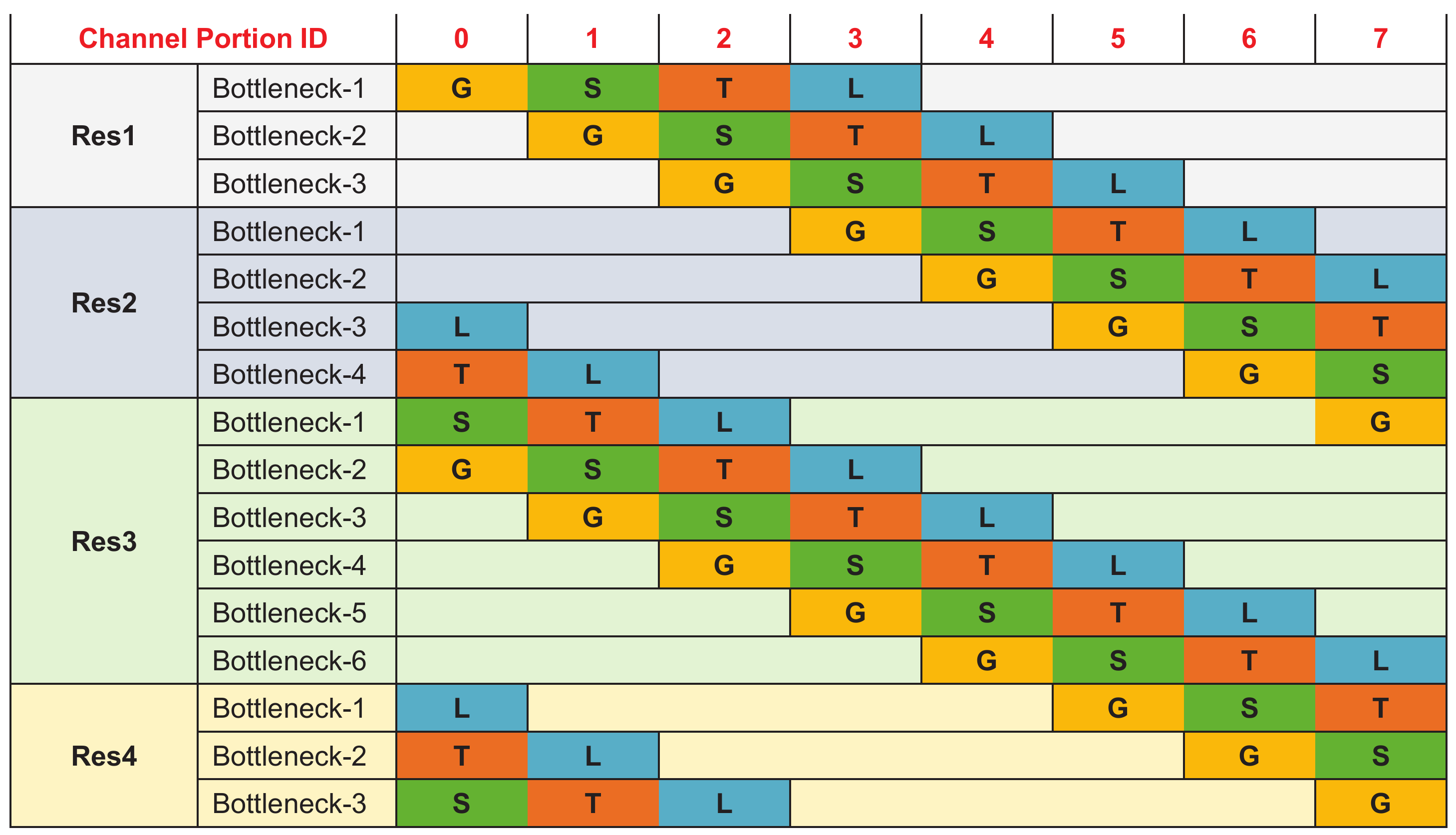}
\vspace{-0.2cm}
\caption{Loop version of GC. The standard 2D ResNet-50 is adopted as the backbone. The partition ratio $p$ is set to $\frac{1}{2}$. G, S, T, L denotes ECal-G, ECal-S, ECal-T and ECal-L units respectively.}
\label{pos_loop}
\vspace{-0.6cm}
\end{figure}

Except the partition ratio $p$ of channels, the position of channel groups could be another discussible variable in GC. To be specific, we empirically investigate a new position setting, i.e., the loop version, as shown in Figure \ref{pos_loop}. Unlike the standard version that remains the channel group positions unchanged in all residual blocks of ResNet, the loop version periodically shifts the channel groups being calibrated block by block with a step of 1. Consequently, the channel groups adjusted by calibrators keep varying in position among residual blocks. In the experiment, we observe different performance tendencies with different backbones. Specifically, we find that there is no significant performance difference between the standard version and the loop version on TSN, TSM and TDN, but the loop version gains a great performance improvement (from 45.6\% to 46.7\% with 8 frames on Something-Something V1 dataset) on GST. This may be because that the convolutional features in TSN/TSM/TDN are entangled together across channels and each subgroup can thus contain similar information (e.g., spatial information in TSN, spatio-temporal information in TSM and short/long-term temporal difference in TDN). Whereas, in GST the spatial and temporal features are separately learnt without any channel interaction between groups. Since the channel groups in GST provide different types of features, keeping on utilizing the same calibrator to adjust fixed channels may be not optimal. For example, when setting $p=\frac{1}{2}$, the spatio-temporal feature group ($\frac{1}{4}C$) outputted by the Conv3d in GST will not participate in the feature calibration in the standard version. If we change $p$ to 1, the spatial feature group ($\frac{3}{4}C$) is fixed to perform the calibration with ECal-G/S/T and the other spatio-temporal feature group is only involved in ECal-L. Differently, the loop version skillfully allows both the two feature groups of GST to achieve feature calibration with all the four ECals. Based on this, we apply the standard version on TSN/TSM/TDN and the loop version on GST in implementation. It is worth mentioning that both the standard and loop versions have the same model complexity.

\vspace{-0.2cm}
\section{Experiment}
\vspace{-0.2cm}
We conduct experiments on different benchmarks, including Something-Something V1\&V2~\cite{goyal2017something,mahdisoltani2018effectiveness} and Kinetics-400 \cite{kay2017kinetics} for video recognition. The four video datasets cover a broad range of activities. Specifically, the \textbf{Something-Something V1\&V2} datasets show 174 fine-grained humans performing pre-defined activities and are more focused on modeling the temporal relationships. The \textbf{Kinetics-400} dataset covers 400 human action classes with less motion variations. 
Due to space limitation, we include the results on EGTEA Gaze+~\cite{li2018eye}, which is a dataset offering first-person videos, Diving48~\cite{li2018resound} with unambiguous dive sequence, and Basketball-8\&Soccer-10 \cite{hao2020compact} with group activities in the supplementary document.

\subsection{Experimental Setup}
\vspace{-0.1cm}
We insert the GC module into four different 2D and 3D ResNets, including TSN, TSM, GST and TDN. Most experiments are based on the backbone of ResNet-50 pretrained on ImageNet~\cite{russakovsky2015imagenet}. Notably, we additionally add a ``BatchNorm'' layer after each convolutional/FC layer in the ECals for TSN, TSM and TDN. We implement GC-Nets in Pytorch and run them on servers with 4$\times$2080Ti or 4/8$\times$3090.

{\bf Training \& Inference}. The \textbf{training} protocol mainly follows the work \cite{wang2016temporal}. Specifically, we use uniform sampling for all datasets. The spatial short side of input frames is resized to 256 maintaining the aspect ratio and then cropped to 224$\times$224. Data augmentation also follows \cite{wang2016temporal}. Training configurations for GC-TSN/TSM/GST are set as follows: a batch-size of 8/10 per GPU, an initial learning rate of 0.01 for 50 epochs and decayed at epoch 20 and 40, the SGD optimizer. GC-TDN follows the training protocol of TDN~\cite{wang2021tdn}. The dropout ratio is set to 0.5.  During the \textbf{inference}, we uniformly sample 8 frames per video and use the 224$\times$224 center crop for performance report in the ablation study. In the final performance comparison, we sample multiple clips per video and take no more than three crops per clip. Specifically, the test protocols are: 2 clips $\times$ 3 crops (224$\times224$) for Something-Something V1\&V2, and 1 clip $\times$ 1 center crop (224$\times224$) for others. We will also specify the sampled frames in the tables.

\subsection{Ablation Study}
We present ablation study to investigate the effect of hyperparameters, including the channel partition ratio $p$, channel position, calibrator variants and backbones, on Something-Something V1 dataset.

\begin{table}[tbp]
		\centering
		\scriptsize
        \begin{tabular}{l|l|c|cc|c}
			\hline
			Backbone &Calibrator &($p$, Channel) &Params &FLOPs &Top-1 (\%)\\
            \midrule[1pt]
            \multirow{18}*{TSN} &--- &--- &23.9M &32.9G &19.7 \\
                    \cline{2-6}
                    &SE3D &--- &26.4M &32.9G &27.8 (+8.1)  \\
                    &GE3D-G &--- &23.9M &32.9G &22.3 (+2.6)  \\
                    &GE3D-C &--- &25.2M &33.3G &44.2 (+24.5)  \\
                    &S3D-G &--- &25.1M &32.9G &28.0 (+8.3)  \\
                    &NLN &--- &31.2M &49.4G &30.3 (+10.6)  \\
                    \cline{2-6}
                    &\multirow{2}*{ECal-G} &($\frac{1}{2}$, $\frac{1}{8}C$) &23.9M &32.9G &26.3 (+6.6) \\
                    & & (1, $\frac{1}{4}C$) &23.9M &32.9G &27.3 (+7.6)  \\
                    \cline{2-6}
                    &\multirow{2}*{ECal-T} &($\frac{1}{2}$, $\frac{1}{8}C$) &23.9M &32.9G &35.9 (+16.2)  \\
                    & &(1, $\frac{1}{4}C$) &24.1M &32.9G &36.4 (+16.7) \\
                    \cline{2-6}
                    &\multirow{2}*{ECal-S} &($\frac{1}{2}$, $\frac{1}{8}C$) &24.0M &32.9G &34.0 (+14.3)  \\
                    & &(1, $\frac{1}{4}C$) &24.6M &33.0G &34.1 (+14.4)  \\
                    \cline{2-6}
                    &\multirow{2}*{ECal-L} &($\frac{1}{2}$, $\frac{1}{8}C$) &23.9M &33.0G &44.8 (+25.1) \\
                    & &(1, $\frac{1}{4}C$) &24.1M &33.2G &44.9 (+25.2) \\
                    \cline{2-6}
                    &\multirow{3}*{GC} &($\frac{1}{2}$, $\frac{1}{2}C$) &24.2M &33.0G &47.1 (+27.4) \\
                    & &(1, $C$) &25.1M &33.3G &47.9 (+28.2) \\
                    & &(1, $C$), loop &25.1M &33.3G &\textbf{48.0 (+28.3)} \\
            \hline
			\multirow{8}*{TSM} &--- &--- &23.9M &32.9G &45.6 \\
			        \cline{2-6}
			        &SE3D &--- &26.4M &32.9G &46.7 (+1.1)  \\
                    &GE3D-G &--- &23.9M &32.9G &45.7 (+0.1)  \\
                    &GE3D-C &--- &25.2M &33.3G &47.0 (+1.4)  \\
                    &S3D-G &--- &25.1M &32.9G &46.8 (+1.2)  \\
                    &NLN &--- &31.2M &49.4G &47.2 (+1.6)  \\
                    \cline{2-6}
                    &\multirow{3}*{GC} &($\frac{1}{2}$, $\frac{1}{2}C$) &24.2M &33.0G &48.7 (+3.1) \\
                    & &(1, $C$) &25.1M &33.3G &\textbf{48.9 (+3.3)} \\
                    & &(1, $C$), loop &25.1M &33.3G &\textbf{48.9 (+3.3)} \\
            \hline
            \multirow{4}*{GST} &--- &--- &21.0M &29.2G &44.4 \\
                    \cline{2-6}
                    &\multirow{3}*{GC} &($\frac{1}{2}$, $\frac{1}{2}C$) &21.4M &29.3G &45.5 (+1.1) \\
                    & &(1, $C$) &22.3M &29.6G &45.6 (+1.2) \\
                    & &(1, $C$), loop &22.3M &29.6G &\textbf{46.7 (+2.3)} \\
            \hline
            \multirow{3}*{TDN} &--- &--- &26.1M &36.0G &52.3 \\
                    \cline{2-6}
                    &\multirow{2}*{GC} &(1, $C$) &27.4M &36.7G &\textbf{53.7 (+1.4)} \\
                    & &(1, $C$), loop &27.4M &36.7G &53.6 (+1.3) \\
            \hline
		\end{tabular}
		\caption{Performance changes using different backbones, calibrators, partition ratio $p$ and channel position on Something-Something V1 dataset. ``Channel'' denotes the number of channels in each calibrated feature group. SE3D is the 3D variant of SE-Net \cite{hu2018squeeze} by replacing the 2D spatial average pooling with the 3D spatio-temporal average pooling. GE3D-G and GE3D-C are two 3D variants of GE-Net \cite{hu2018gather}, where GE3D-G adopts global average pooling and GE3D-C employs 3D depthwise convolution. Their architectures can be found in Appendix. NLN denotes the nonlocal~\cite{wang2018non} module.}
		\label{tab:res_ab}
		\vspace{-0.5cm}
		\end{table}

{\bf $p$ and calibrators}. We first compare different ECals on TSN with $p=\frac{1}{2},1$. The four types of ECals are designed to calibrate video feature with different axial context concerns. And the channel partition ratio $p$ is introduced to control the number of channels to be calibrated by ECals. As shown in Table \ref{tab:res_ab}, we observe that ECal variants, regardless of their types, consistently improve the recognition performance of the backbone TSN, indicating their effectiveness. Although varying the value of $p$ from $\frac{1}{2}$ to 1 will result in slight increase of model size and computational cost, the performance boost is noticeable (e.g., 26.3\%$\rightarrow$27.3\% for ECal-G and 35.9\%$\rightarrow$36.4\% for ECal-T). 

{\bf Channel position and backbones}. Secondly, we test both the standard and loop GC versions on the four backbones. Here, we also set $p=\frac{1}{2},1$. Table \ref{tab:res_ab} shows their results. Compared to the single calibrator, the GC module, which combines the four ECals in parallel, achieves much better performance on TSN. The GC-TSM, GC-GST and GC-TDN also gain significant performance improvement (45.6\%$\rightarrow$48.9\% for TSM, 44.4\%$\rightarrow$46.7\% for GST, 52.3\%$\rightarrow$53.7\% for TDN) to their original backbones. Consistently, the models with larger $p=1$ outperform their counterparts with $p=\frac{1}{2}$. Based on the above results, we fix $p=1$ for the GC-Nets in this work. For the channel position, we observe different performance tendencies on the four backbones, i.e., the result of loop version is clearly better than the standard version on GST, and their performances are about the same on TSN, TSM and TDN. As analysed in Section \ref{channelposition}, this is because that feature channels in TSN, TSM and TDN are entangled together during the feature learning while the group convolution method GST separately models the spatial and temporal features. 

{\bf Comparison with other calibrators}. Thirdly, we integrate the 3D variants of SE-Net~\cite{hu2018squeeze} and GE-Net~\cite{hu2018gather}, i.e., SE3D and GE3D-G/C, S3D-G and NLN, into the TSN and TSM backbones. Their hyperparameters are set as the same to their original papers. The NLN-Nets follows the implementation of \cite{lin2019tsm}. From Table \ref{tab:res_ab}, we can find that our GC module far outstrips SE3D, GE3D-G and S3D-G which only consider the global context and the pairwise self-attention NLN when using TSN as backbone. Since GE3D-C uses three 3D depthwise convolution layers to model local spatio-temporal context, relatively good performance (44.2\% Top-1 accuracy) is attained on TSN but still lower than our GC (47.9\%). On TSM, our GC can outperform the five other calibrators by the margins of 1.7\%-3.2\%. Moreover, compared to the self-attention NLN module that results in 31\% extra parameters and 50\% extra FLOPs to the backbones, our GC only introduces as low as 5\% extra parameters and 1.2\% extra FLOPs.

\begin{figure}[]
\centering
\includegraphics[width=0.45\textwidth]{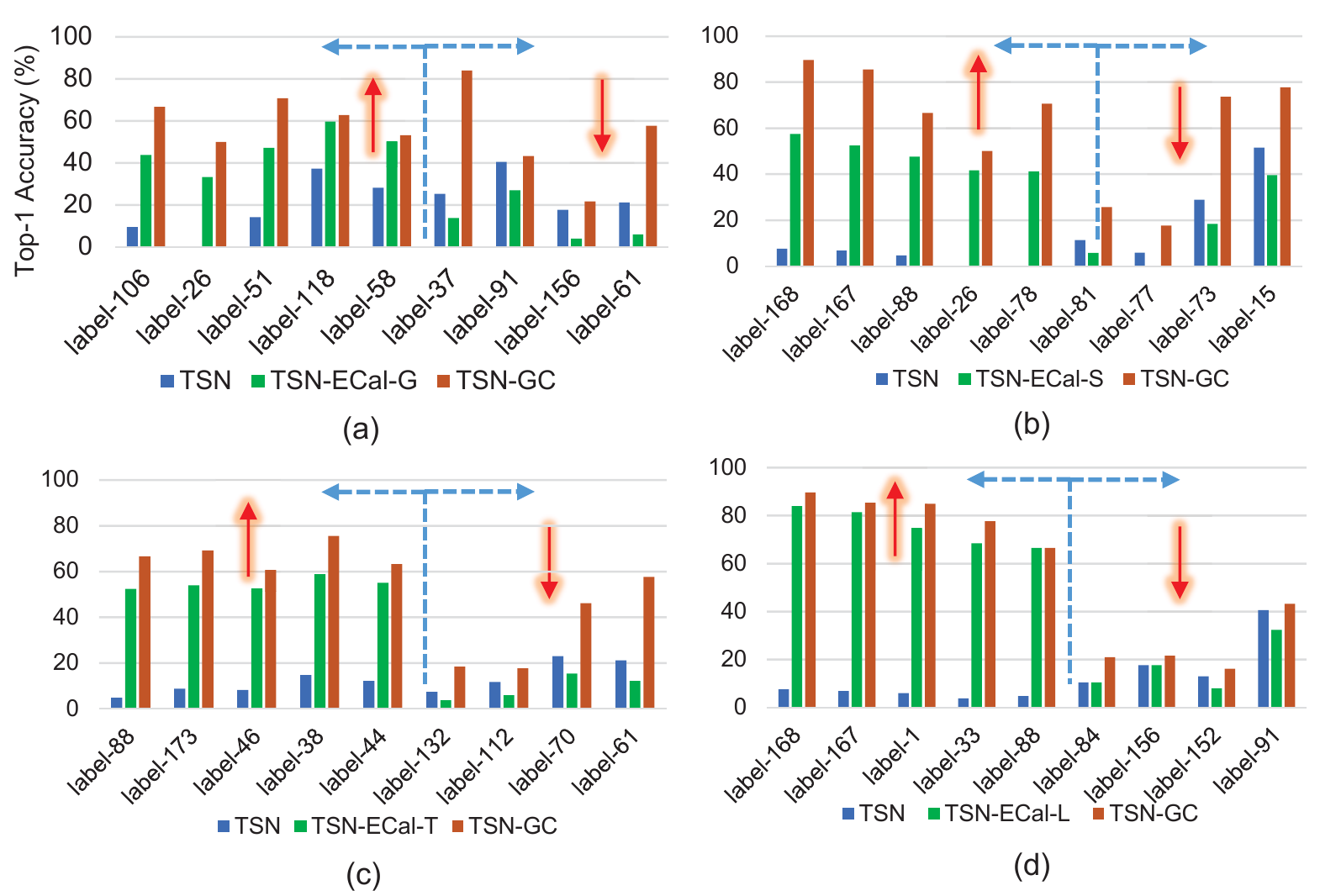}
\vspace{-0.4cm}
\caption{Per-category accuracy comparison for TSN, TSN-ECals and GC-TSN over several selected activity categories on Something-Something V1 dataset (validation set). In each subfigure, we show 5 action categories that are improved most by TSN-ECal and another 4 activity categories that are being degraded by TSN-ECal but improved by GC-TSN. See the supplementary material for the label details of these selected activities.}
\label{figure:bars}
\vspace{-0.6cm}
\end{figure}

\subsection{Example Demonstration}
\vspace{-0.1cm}
We show the per-category results of TSN-ECal variants and GC-TSN to understand the impact of axial contexts on different kinds of video activities in Figure \ref{figure:bars}. Specifically, ECal-G can boost the recognition of activities that need global contexts, e.g., ``\textit{label-106: Putting something in front of something}'' in Figure \ref{figure:bars}(a). ECal-S focuses on enhancing the feature spatial-wisely with the context aggregated along temporal dimension and thus can improve the performance for activities that require more spatial than temporal information, for example ``\textit{label-26: Lifting a surface with something on it but not enough for it to slide down}'' in \ref{figure:bars}(b). TSN-ECal-T and TSN-ECal-L, which naturally calibrate the feature globally and locally along temporal dimension respectively, can significantly improve the recognition performance for activities that require temporal reasoning, as shown in the first 5 categories in figures \ref{figure:bars}(c) and (d). However, those failure cases in the four subfigures (last 4 terms) provide evidence that one kind of axial contexts is not suitable for all activity categories. For example, TSN-ECal-G fails to model the activities that involve rich spatial-temporal interactions between objects, e.g., ``\textit{label-61: Pouring something into something until it overflows}'', and TSN-ECal-S under-performs on ``\textit{label-73: Pretending to put something into something}'' and ``\textit{label-81: Pretending to squeeze something}'' which require strong temporal reasoning. Encouragingly, the GC module that aggregates all the four ECals indeed alleviates these shortcoming of individual ECal, leading to  performance improvement for a variety of activities.

\subsection{Comparison with the State-of-the-Arts}
We compare GC-Nets with state-of-the-art networks in this section. The result comparison follows the same protocol of using RGB frames as input and adopting ResNet50 unless otherwise specified.

\begin{table}[t]
		\centering
		\scriptsize
		\begin{tabular}{l|c|c|c|cc}
			\hline
			Method &Params &\#Frame &FLOPs$\times$Clips &Top-1 &Top-5 \\
            \midrule[1pt]
            CorrNet \cite{wang2020video} &---  &32 &115.0G$\times$10  &49.3 &---  \\
            TIN \cite{shao2020temporal} &24.6M$\times$2  &8+16 &101G$\times$1 &49.6 &78.3  \\
            V4D \cite{zhang2020v4d} &---  &8$\times$4 &167.6G$\times$30 &50.4 &---  \\
            SmallBig \cite{li2020smallbignet} &---  &8+16 &157G$\times$1 &50.4 &80.5  \\
            TANet~\cite{liu2020tam} &25.1M$\times$2  &8+16 &99.0G$\times$1 &50.6 &79.3  \\
            STM \cite{jiang2019stm} &24.0M  &16 &33.3G$\times$30  &50.7 &80.4  \\
            TEA \cite{li2020tea}   &---  &16 &70.0G$\times$30 &52.3 &81.9  \\
            TEINet \cite{liu2020teinet} &30.4M$\times$2  &8+16 &99.0G$\times$1 &52.5 &---  \\
            RNL-TSM \cite{huang2021region} &35.5M$\times$2  &8+16 &123.5G$\times$2 &52.7 &---  \\
            \Xcline{1-6}{0.7pt}
            GST* \cite{luo2019grouped} &21.0M &16 &58.4G$\times$2  &46.2 &75.0 \\
            GST* \cite{luo2019grouped} &21.0M$\times$2 &8+16 &87.6G$\times$2  &48.6 &78.3 \\
            TSM \cite{lin2019tsm} &23.9M  &16 &65.8G$\times$2  &48.4 &78.1  \\
            TSM \cite{lin2019tsm} &23.9M$\times$2  &8+16 &98.7G$\times$2   &50.3 &79.3 \\
            TDN \cite{wang2021tdn} &26.1M  &16 &72.0G$\times$1 &53.9 &82.1  \\
            TDN \cite{wang2021tdn} &26.1M$\times$2  &8+16 &108.0G$\times$1 &55.1 &82.9  \\
            \Xcline{1-6}{0.7pt}
            \textbf{GC-GST} &22.3M  &8 &29.6G$\times$2 &48.8 &78.5  \\
            \textbf{GC-GST} &22.3M  &16 &59.1G$\times$2  &50.4 &79.4  \\
            \textbf{GC-GST} &22.3M$\times$2  &8+16 &88.7G$\times$2  &52.5 &81.3  \\
            \hline
            \textbf{GC-TSN} &25.1M  &8 &33.3G$\times$2 &49.7 &78.2 \\

            \textbf{GC-TSN} &25.1M  &16 &66.5G$\times$2  &51.3 &80.0  \\
            \textbf{GC-TSN} &25.1M$\times$2  &8+16 &99.8G$\times$2  &53.7 &81.8 \\
            \hline
            
            \textbf{GC-TSM} &25.1M  &8 &33.3G$\times$2 &51.1 &79.4  \\

            
            \textbf{GC-TSM} &25.1M  &16 &66.5G$\times$2  &53.1 &81.2  \\
            
            \textbf{GC-TSM} &25.1M$\times$2  &8+16 &99.8G$\times$2  &55.0 &82.6  \\
            \textbf{GC-TSM} &25.1M$\times$2  &8+16 &99.8G$\times$6  &55.3 &82.7  \\
            \hline
            \textbf{GC-TDN} &27.4M  &8 &36.7G$\times$1 &53.7 &82.2  \\
            \textbf{GC-TDN} &27.4M  &16 &73.4G$\times$1 &55.0 &82.3  \\
            \textbf{GC-TDN} &27.4M$\times$2  &8+16 &110.1G$\times$1 &\textbf{56.4} &\textbf{84.0} \\
            \hline
		\end{tabular}
		\vspace{-0.3cm}
		\caption{Comparison of performance on Something-Something V1 dataset. ``*'' indicates that the result is obtained by ourselves.}
		\label{tab:res_somev1}
		\vspace{-0.5cm}
\end{table}

\begin{table}[!t]
		\centering
		\scriptsize
		\begin{tabular}{l|c|c|c|cc}
			\hline
			Method &Params &\#Frame &FLOPs$\times$Clips &Top-1 &Top-5 \\
            \midrule[1pt]
            TIN \cite{shao2020temporal} &24.6M  &16 &67.0G$\times$1 &60.1 &86.4 \\
            RubiksNet \cite{fan2020rubiksnet} &8.5M  &8 &15.8G$\times$2  &61.7 &87.3  \\
            TSM+TPN \cite{yang2020temporal} &---  &8 &33.0G$\times$1 &62.0 &---  \\
            SlowFast \cite{feichtenhofer2019slowfast} &32.9M  &4+32 &65.7G$\times$6  &61.9 &87.0 \\
            SlowFast(R101) \cite{feichtenhofer2019slowfast} &53.3M  &8+32 &106G$\times$6  &63.1 &87.6 \\
            SmallBig \cite{li2020smallbignet} &---  &16 &114.0G$\times$6 &63.8 &88.9  \\
            STM \cite{jiang2019stm} &24.0M  &16 &33.3G$\times$30 &64.2 &89.8  \\
            TEA \cite{li2020tea}   &---  &16 &70.0G$\times$30 &65.1 &89.9  \\
            TEINet \cite{liu2020teinet} &30.4M$\times$2  &8+16  &99.0G$\times$1 &65.5 &89.8  \\
            TANet~\cite{liu2020tam} &25.1M$\times$2  &8+16 &99.0G$\times$6 &66.0 &90.1  \\
            \Xcline{1-6}{0.7pt}
            TimeSformer-HR~\cite{bertasius2021space} &121.4M  &16 &1703G$\times$3  &62.5 &--- \\
            ViViT-L~\cite{arnab2021vivit} &352.1M &32 &903G$\times$4  &65.4 &89.8 \\
            MViT-B~\cite{fan2021multiscale} &36.6M  &64 &455G$\times$3  &67.7 &90.9 \\
            Video-Swin-B~\cite{liu2021video} &88.8M  &16 &321G$\times$3  &\textbf{69.6} &\textbf{92.7} \\
            \Xcline{1-6}{0.7pt}
            TSN \cite{zhou2018temporal} from \cite{lin2019tsm}  &23.9M  &8 &32.9G$\times$1 &30.0 &60.5 \\
            GST* \cite{luo2019grouped} &21.0M &8 &29.2G$\times$2  &59.8 &86.3 \\
            GST* \cite{luo2019grouped} &21.0M &16 &58.4G$\times$2  &61.7 &87.2 \\
            GST* \cite{luo2019grouped} &21.0M$\times$2 &8+16 &87.6G$\times$2  &63.1 &88.3 \\
            TSM \cite{lin2019tsm} &23.9M  &8 &32.9G$\times$2   &61.2 &87.1  \\
            TSM \cite{lin2019tsm} &23.9M  &16 &65.8G$\times$2  &63.1 &88.2  \\
            TSM \cite{lin2019tsm} &23.9M$\times$2  &8+16 &98.7G$\times$2  &64.3 &89.0 \\
            TDN \cite{wang2021tdn} &26.1M  &8 &36.0G$\times$1 &64.0 &88.8  \\
            TDN \cite{wang2021tdn} &26.1M  &16 &72.0G$\times$1 &65.3 &89.5  \\
            TDN \cite{wang2021tdn} &26.1M$\times$2  &8+16 &108G$\times$1 &67.0 &90.3  \\
            \Xcline{1-6}{0.7pt}
            \textbf{GC-GST} &22.3M  &8 &29.6G$\times$2 &61.9 &87.8  \\
            \textbf{GC-GST} &22.3M  &16 &59.1G$\times$2  &63.3 &88.5  \\
            \textbf{GC-GST} &22.3M$\times$2  &8+16  &88.7G$\times$2  &65.0 &89.5  \\
            \hline
            \textbf{GC-TSN} &25.1M  &8  &33.3G$\times$2 &62.4 &87.9 \\

            \textbf{GC-TSN} &25.1M  &16  &66.5G$\times$2  &64.8 &89.4  \\
            \textbf{GC-TSN} &25.1M  &8+16  &99.8G$\times$2  &66.3 &90.3 \\
            \hline
            
            \textbf{GC-TSM} &25.1M  &8  &33.3G$\times$2 &63.0 &88.4  \\
            
            \textbf{GC-TSM} &25.1M  &16  &66.5G$\times$2  &64.9 &89.7  \\
            
            \textbf{GC-TSM} &25.1M$\times$2  &8+16 &99.8G$\times$2  &66.7 &90.6 \\
            \textbf{GC-TSM} &25.1M$\times$2  &8+16 &99.8G$\times$6  &67.5 &90.9 \\
            
            \hline
            \textbf{GC-TDN} &27.4M  &8  &36.7G$\times$1 &64.9 &89.7  \\
            \textbf{GC-TDN} &27.4M  &16  &73.4G$\times$1 &65.9 &90.0  \\
            \textbf{GC-TDN} &27.4M$\times$2  &8+16  &110.1G$\times$1 &\textbf{67.8} &\textbf{91.2}  \\
            \hline
		\end{tabular}
		\caption{Comparison of performance on Something-Something V2 dataset. ``*'' indicates that the result is obtained by ourselves.}
		\label{tab:res_somev2}
		\vspace{-0.5cm}
\end{table}

{\bf Something-Something V1 \& V2}. A comprehensive comparison between our GC-Nets and SOTAs on Something-Something V1\&V2  datasets are presented. Tables \ref{tab:res_somev1} and \ref{tab:res_somev2} list the comparison in terms of Top-1/5 accuracy, FLOPs and model complexity. GC-TDN achieves the highest Top-1 accuracies of 56.4\% and 67.8\% with (8+16) frames $\times$ 1 clip on Something-Something V1\&V2, respectively, which outperform all the CNN-based SOTAs by large margins (1.3\%-36.7\% for V1 and 0.8\%-37.8\% for V2). Moreover, all GC-Nets, including GC-GST, GC-TSN, GC-TSM and GC-TDN, consistently outperform their backbone networks with significant performance gains, demonstrating the capacity of GC module in recognizing diverse activities and the strong versatility against various deep video networks. For example, GC-TSN boosts the original TSN model with an absolute improvements of 28.2\% (19.7\%$\rightarrow$47.9\%) on V1 and 32.4\% (30.0\%$\rightarrow$62.4\%) on V2 with the same 8-frame input. Equipping TSN, which is an image-based CNN, with GC empowers the modeling of temporal relationship between objects on V1\&V2. The GC module can also improve the more advanced TDN by 1.3\% on V1 and 0.8\% on V2 with the same 8+16 frames. This demonstrates that the axial contexts modeled by GC can work cooperatively with the temporal difference contexts used by TDN. Compared to the more sophisticated Transformer-based models like MViT and Video-Swin, the GC-TSM and GC-TDN obtain lower Top-1 accuracies. The performance is compensated by lower computational cost and model complexity. GC-TDN requires 110.1G FLOPs, which is about 11.4 times cheaper than MViT-B (1,365G FLOPs) and 7.7 times lower than Video-Swin-B (963G FLOPs).

{\bf Kinetcis-400}. We report the results of GC-TSN with 8 frames and GC-TSM/TD with 8 and 16 frames respectively, in Table \ref{tab:res_kinetic}. Firstly, the GC-TSN/TSM/TDN improve the Top1 accuracy upon TSN/TSM/TDN by 4.6\%/2.0\%/1.2\% under the same input, respectively. They are much significant in terms of the data scale of Kinetics-400 dataset. Secondly, GC-TDN, with 16$\times$30 clips as input, achieves the 79.6\% Top1 accuracy, which is the highest one among the competing methods. This result is much better than other models equipped with feature contextualization techniques, such as  Nonlocal-I3D, S3D-G and TEA, which further demonstrates the superior performance of the proposed GC module. 

\begin{table}[htbp]
		\centering
		\scriptsize
		\begin{tabular}{l|c|c|c|cc}
			\hline
			Model &Params &\#Frame &FLOPs$\times$Clips &Top1 &Top5 \\
            \midrule[1pt]
            I3D (InceptionV1) \cite{carreira2017quo} & --- &64 &---  &72.1 &90.3 \\
            Nonlocal-I3D \cite{wang2018non} &35.3M  &32 &282G$\times$10  &74.9 &91.6 \\
            S3D-G (InceptionV1) \cite{xie2018rethinking} &--- &64 &71.4G$\times$30 &74.7 &\textbf{93.4} \\
            TEA \cite{li2020tea} &--- & 16 &70G$\times$30 &76.1 &92.5 \\
            TEINet \cite{liu2020teinet} &30.8M & 16 &66G$\times$30 &76.2 &92.5 \\
            TANet \cite{liu2020tam} &25.6M &16 &86G$\times$12 &76.9 &92.9 \\
            SmallBig \cite{li2020smallbignet} &--- &8 &57G$\times$30 &76.3  &92.5 \\
            SlowFast(8$\times$8) \cite{feichtenhofer2019slowfast} &32.9M &8+32 &65.7G$\times$30 &77.0 &92.6 \\
            X3D-L \cite{feichtenhofer2020x3d} &6.1M & 16 &24.8G$\times$30 &77.5 &92.9 \\
            \hline
            TSN \cite{zhou2018temporal} &24.3M &8 &32.9G$\times$10clip  &70.6 &89.2 \\
            TSM \cite{lin2019tsm} &24.3M  &16 &66.0G$\times$10 &74.7 &91.4 \\
            TDN \cite{wang2021tdn} &26.6M &8+16 &108.0G$\times$30 &78.4  &93.6 \\
            \hline
            \textbf{GC-TSN} &25.6M &8 &33.3G$\times$10 &75.2  &92.1 \\
            \textbf{GC-TSM} &25.6M &8 &33.3G$\times$10 &75.4  &91.9 \\
            \textbf{GC-TSM} &25.6M &16 &66.6G$\times$10 &76.7  &92.9 \\
            \textbf{GC-TSM} &25.6M &16 &66.6G$\times$30 &77.1  &92.9 \\
            \textbf{GC-TDN} &27.4M &8 &36.7G$\times$30 &77.3  &93.2 \\
            \textbf{GC-TDN} &27.4M &16 &73.4G$\times$30 &78.8  &93.8 \\
            \textbf{GC-TDN} &27.4M &8+16 &110.1G$\times$30 &\textbf{79.6}  &94.1 \\
            \hline
		\end{tabular}
		\caption{Comparison of performance on Kinetics-400 dataset.}
		\vspace{-0.3cm}
		\label{tab:res_kinetic}
\end{table}

\section{Conclusion}
\vspace{-0.2cm}
We have presented the regime of group contextualization, which aims at deriving robust representations generic to various video activities by calibrating plain features computed from off-the-shelf networks with multiple contexts. The family of element-wise calibrators is designed to work on different grouped feature channels independently. The group operation results in a much lower computation cost increase (5.3\%/1.3\% extra parameters/FLOPs) and substantial performance improvements (0.4\%-32.4\%) to backbones. More surprisingly, when GC module is integrated into the 2D spatial TSN model, GC-TSN achieves absolute 28.2\%/32.4\% performance improvements on Something-Something V1/V2 and even performs much better than the advanced 3D spatio-temporal GST and TSM models.~We conclude that since the videos in Something-Something datasets contain rich global/local human-object interactions, GC module that explores various global/local spatial/temporal axial contexts to calibrate the original feature exhibits excellent performance. Similar results are also observed from the other datasets (e.g., Diving and Kitchen Activities). Moreover, compared to the other feature calibration methods, such as SE3D, GE3D, S3D-G, TEA and TANet that only use a single context, GC-Nets consistently achieve better performances, which further proves the feasibility and advantages of the proposed group contextualization. The significant performance improvement of GC-TDN further demonstrates that our GC can also work together with the other temporal difference context (TDN).

\vspace{-0.2cm}
\section*{Acknowledgements}
\vspace{-0.2cm}
The work was supported in part by the National Natural Science Foundation of China (No. 62101524), by the National Key Research and Development Program of China under Grants 2020YFB1406703, and by the Singapore Ministry of Education (MOE) Academic Research Fund (AcRF) Tier 1 grant.

{\small
\bibliographystyle{ieee_fullname}
\bibliography{refer}

\begin{thebibliography}{10}\itemsep=-1pt

\bibitem{arnab2021vivit}
Anurag Arnab, Mostafa Dehghani, Georg Heigold, Chen Sun, Mario Lu{\v{c}}i{\'c},
  and Cordelia Schmid.
\newblock Vivit: A video vision transformer.
\newblock {\em arXiv preprint arXiv:2103.15691}, 2021.

\bibitem{bertasius2021space}
Gedas Bertasius, Heng Wang, and Lorenzo Torresani.
\newblock Is space-time attention all you need for video understanding?
\newblock {\em arXiv preprint arXiv:2102.05095}, 2021.

\bibitem{carreira2017quo}
Joao Carreira and Andrew Zisserman.
\newblock Quo vadis, action recognition? a new model and the kinetics dataset.
\newblock In {\em proceedings of the IEEE Conference on Computer Vision and
  Pattern Recognition}, pages 6299--6308, 2017.

\bibitem{fan2021multiscale}
Haoqi Fan, Bo Xiong, Karttikeya Mangalam, Yanghao Li, Zhicheng Yan, Jitendra
  Malik, and Christoph Feichtenhofer.
\newblock Multiscale vision transformers.
\newblock {\em arXiv preprint arXiv:2104.11227}, 2021.

\bibitem{fan2020rubiksnet}
Linxi Fan, Shyamal Buch, Guanzhi Wang, Ryan Cao, Yuke Zhu, Juan~Carlos Niebles,
  and Li Fei-Fei.
\newblock Rubiksnet: Learnable 3d-shift for efficient video action recognition.
\newblock In {\em European Conference on Computer Vision}, pages 505--521.
  Springer, 2020.

\bibitem{feichtenhofer2020x3d}
Christoph Feichtenhofer.
\newblock X3d: Expanding architectures for efficient video recognition.
\newblock In {\em Proceedings of the IEEE/CVF Conference on Computer Vision and
  Pattern Recognition}, pages 203--213, 2020.

\bibitem{feichtenhofer2019slowfast}
Christoph Feichtenhofer, Haoqi Fan, Jitendra Malik, and Kaiming He.
\newblock Slowfast networks for video recognition.
\newblock In {\em Proceedings of the IEEE/CVF International Conference on
  Computer Vision}, pages 6202--6211, 2019.

\bibitem{fu2019dual}
Jun Fu, Jing Liu, Haijie Tian, Yong Li, Yongjun Bao, Zhiwei Fang, and Hanqing
  Lu.
\newblock Dual attention network for scene segmentation.
\newblock In {\em Proceedings of the IEEE/CVF Conference on Computer Vision and
  Pattern Recognition}, pages 3146--3154, 2019.

\bibitem{goyal2017something}
Raghav Goyal, Samira Ebrahimi~Kahou, Vincent Michalski, Joanna Materzynska,
  Susanne Westphal, Heuna Kim, Valentin Haenel, Ingo Fruend, Peter Yianilos,
  Moritz Mueller-Freitag, et~al.
\newblock The" something something" video database for learning and evaluating
  visual common sense.
\newblock In {\em Proceedings of the IEEE International Conference on Computer
  Vision}, pages 5842--5850, 2017.

\bibitem{guo2021ssan}
Xudong Guo, Xun Guo, and Yan Lu.
\newblock Ssan: Separable self-attention network for video representation
  learning.
\newblock In {\em Proceedings of the IEEE/CVF Conference on Computer Vision and
  Pattern Recognition}, pages 12618--12627, 2021.

\bibitem{guo2020visual}
Yutian Guo, Jingjing Chen, Hao Zhang, and Yu-Gang Jiang.
\newblock Visual relations augmented cross-modal retrieval.
\newblock In {\em Proceedings of the 2020 International Conference on
  Multimedia Retrieval}, pages 9--15, 2020.

\bibitem{han2021fine}
Ning Han, Jingjing Chen, Guangyi Xiao, Hao Zhang, Yawen Zeng, and Hao Chen.
\newblock Fine-grained cross-modal alignment network for text-video retrieval.
\newblock In {\em Proceedings of the 29th ACM International Conference on
  Multimedia}, pages 3826--3834, 2021.

\bibitem{hao2020compact}
Yanbin Hao, Hao Zhang, Chong-Wah Ngo, Qiang Liu, and Xiaojun Hu.
\newblock Compact bilinear augmented query structured attention for sport
  highlights classification.
\newblock In {\em Proceedings of the 28th ACM International Conference on
  Multimedia}, pages 628--636, 2020.

\bibitem{hara2018can}
Kensho Hara, Hirokatsu Kataoka, and Yutaka Satoh.
\newblock Can spatiotemporal 3d cnns retrace the history of 2d cnns and
  imagenet?
\newblock In {\em Proceedings of the IEEE conference on Computer Vision and
  Pattern Recognition}, pages 6546--6555, 2018.

\bibitem{hu2018gather}
Jie Hu, Li Shen, Samuel Albanie, Gang Sun, and Andrea Vedaldi.
\newblock Gather-excite: exploiting feature context in convolutional neural
  networks.
\newblock In {\em Proceedings of the 32nd International Conference on Neural
  Information Processing Systems}, pages 9423--9433, 2018.

\bibitem{hu2018squeeze}
Jie Hu, Li Shen, and Gang Sun.
\newblock Squeeze-and-excitation networks.
\newblock In {\em Proceedings of the IEEE conference on computer vision and
  pattern recognition}, pages 7132--7141, 2018.

\bibitem{huang2021region}
Guoxi Huang and Adrian~G Bors.
\newblock Region-based non-local operation for video classification.
\newblock In {\em 2020 25th International Conference on Pattern Recognition
  (ICPR)}, pages 10010--10017. IEEE, 2021.

\bibitem{huang2017densely}
Gao Huang, Zhuang Liu, Laurens Van Der~Maaten, and Kilian~Q Weinberger.
\newblock Densely connected convolutional networks.
\newblock In {\em Proceedings of the IEEE conference on computer vision and
  pattern recognition}, pages 4700--4708, 2017.

\bibitem{jiang2019stm}
Boyuan Jiang, MengMeng Wang, Weihao Gan, Wei Wu, and Junjie Yan.
\newblock Stm: Spatiotemporal and motion encoding for action recognition.
\newblock In {\em Proceedings of the IEEE/CVF International Conference on
  Computer Vision}, pages 2000--2009, 2019.

\bibitem{kay2017kinetics}
Will Kay, Joao Carreira, Karen Simonyan, Brian Zhang, Chloe Hillier, Sudheendra
  Vijayanarasimhan, Fabio Viola, Tim Green, Trevor Back, Paul Natsev, et~al.
\newblock The kinetics human action video dataset.
\newblock {\em arXiv preprint arXiv:1705.06950}, 2017.

\bibitem{krizhevsky2012imagenet}
Alex Krizhevsky, Ilya Sutskever, and Geoffrey~E Hinton.
\newblock Imagenet classification with deep convolutional neural networks.
\newblock {\em Advances in neural information processing systems},
  25:1097--1105, 2012.

\bibitem{li2020smallbignet}
Xianhang Li, Yali Wang, Zhipeng Zhou, and Yu Qiao.
\newblock Smallbignet: Integrating core and contextual views for video
  classification.
\newblock In {\em Proceedings of the IEEE/CVF Conference on Computer Vision and
  Pattern Recognition}, pages 1092--1101, 2020.

\bibitem{li2020tea}
Yan Li, Bin Ji, Xintian Shi, Jianguo Zhang, Bin Kang, and Limin Wang.
\newblock Tea: Temporal excitation and aggregation for action recognition.
\newblock In {\em Proceedings of the IEEE/CVF Conference on Computer Vision and
  Pattern Recognition}, pages 909--918, 2020.

\bibitem{li2018resound}
Yingwei Li, Yi Li, and Nuno Vasconcelos.
\newblock Resound: Towards action recognition without representation bias.
\newblock In {\em Proceedings of the European Conference on Computer Vision
  (ECCV)}, pages 513--528, 2018.

\bibitem{li2018eye}
Yin Li, Miao Liu, and James~M Rehg.
\newblock In the eye of beholder: Joint learning of gaze and actions in first
  person video.
\newblock In {\em Proceedings of the European Conference on Computer Vision
  (ECCV)}, pages 619--635, 2018.

\bibitem{lin2019tsm}
Ji Lin, Chuang Gan, and Song Han.
\newblock Tsm: Temporal shift module for efficient video understanding.
\newblock In {\em Proceedings of the IEEE/CVF International Conference on
  Computer Vision}, pages 7083--7093, 2019.

\bibitem{liu2020teinet}
Zhaoyang Liu, Donghao Luo, Yabiao Wang, Limin Wang, Ying Tai, Chengjie Wang,
  Jilin Li, Feiyue Huang, and Tong Lu.
\newblock Teinet: Towards an efficient architecture for video recognition.
\newblock In {\em Proceedings of the AAAI Conference on Artificial
  Intelligence}, volume~34, pages 11669--11676, 2020.

\bibitem{liu2021video}
Ze Liu, Jia Ning, Yue Cao, Yixuan Wei, Zheng Zhang, Stephen Lin, and Han Hu.
\newblock Video swin transformer.
\newblock {\em arXiv preprint arXiv:2106.13230}, 2021.

\bibitem{liu2020tam}
Zhaoyang Liu, Limin Wang, Wayne Wu, Chen Qian, and Tong Lu.
\newblock Tam: Temporal adaptive module for video recognition.
\newblock {\em arXiv preprint arXiv:2005.06803}, 2020.

\bibitem{luo2019grouped}
Chenxu Luo and Alan~L Yuille.
\newblock Grouped spatial-temporal aggregation for efficient action
  recognition.
\newblock In {\em Proceedings of the IEEE/CVF International Conference on
  Computer Vision}, pages 5512--5521, 2019.

\bibitem{mahdisoltani2018effectiveness}
Farzaneh Mahdisoltani, Guillaume Berger, Waseem Gharbieh, David Fleet, and
  Roland Memisevic.
\newblock On the effectiveness of task granularity for transfer learning.
\newblock {\em arXiv preprint arXiv:1804.09235}, 2018.

\bibitem{misra2021rotate}
Diganta Misra, Trikay Nalamada, Ajay~Uppili Arasanipalai, and Qibin Hou.
\newblock Rotate to attend: Convolutional triplet attention module.
\newblock In {\em Proceedings of the IEEE/CVF Winter Conference on Applications
  of Computer Vision}, pages 3139--3148, 2021.

\bibitem{qiu2017learning}
Zhaofan Qiu, Ting Yao, and Tao Mei.
\newblock Learning spatio-temporal representation with pseudo-3d residual
  networks.
\newblock In {\em proceedings of the IEEE International Conference on Computer
  Vision}, pages 5533--5541, 2017.

\bibitem{russakovsky2015imagenet}
Olga Russakovsky, Jia Deng, Hao Su, Jonathan Krause, Sanjeev Satheesh, Sean Ma,
  Zhiheng Huang, Andrej Karpathy, Aditya Khosla, Michael Bernstein, et~al.
\newblock Imagenet large scale visual recognition challenge.
\newblock {\em International journal of computer vision}, 115(3):211--252,
  2015.

\bibitem{selvaraju2017grad}
Ramprasaath~R Selvaraju, Michael Cogswell, Abhishek Das, Ramakrishna Vedantam,
  Devi Parikh, and Dhruv Batra.
\newblock Grad-cam: Visual explanations from deep networks via gradient-based
  localization.
\newblock In {\em Proceedings of the IEEE international conference on computer
  vision}, pages 618--626, 2017.

\bibitem{shao2020temporal}
Hao Shao, Shengju Qian, and Yu Liu.
\newblock Temporal interlacing network.
\newblock In {\em Proceedings of the AAAI Conference on Artificial
  Intelligence}, volume~34, pages 11966--11973, 2020.

\bibitem{sudhakaran2020gate}
Swathikiran Sudhakaran, Sergio Escalera, and Oswald Lanz.
\newblock Gate-shift networks for video action recognition.
\newblock In {\em Proceedings of the IEEE/CVF Conference on Computer Vision and
  Pattern Recognition}, pages 1102--1111, 2020.

\bibitem{sudhakaran2018attention}
Swathikiran Sudhakaran and Oswald Lanz.
\newblock Attention is all we need: Nailing down object-centric attention for
  egocentric activity recognition.
\newblock {\em arXiv preprint arXiv:1807.11794}, 2018.

\bibitem{tan2021vimpac}
Hao Tan, Jie Lei, Thomas Wolf, and Mohit Bansal.
\newblock Vimpac: Video pre-training via masked token prediction and
  contrastive learning.
\newblock {\em arXiv preprint arXiv:2106.11250}, 2021.

\bibitem{tan2021selective}
Yi Tan, Yanbin Hao, Xiangnan He, Yinwei Wei, and Xun Yang.
\newblock Selective dependency aggregation for action classification.
\newblock In {\em Proceedings of the 29th ACM International Conference on
  Multimedia}, pages 592--601, 2021.

\bibitem{tran2015learning}
Du Tran, Lubomir Bourdev, Rob Fergus, Lorenzo Torresani, and Manohar Paluri.
\newblock Learning spatiotemporal features with 3d convolutional networks.
\newblock In {\em Proceedings of the IEEE international conference on computer
  vision}, pages 4489--4497, 2015.

\bibitem{tran2019video}
Du Tran, Heng Wang, Lorenzo Torresani, and Matt Feiszli.
\newblock Video classification with channel-separated convolutional networks.
\newblock In {\em Proceedings of the IEEE/CVF International Conference on
  Computer Vision}, pages 5552--5561, 2019.

\bibitem{wang2017residual}
Fei Wang, Mengqing Jiang, Chen Qian, Shuo Yang, Cheng Li, Honggang Zhang,
  Xiaogang Wang, and Xiaoou Tang.
\newblock Residual attention network for image classification.
\newblock In {\em Proceedings of the IEEE conference on computer vision and
  pattern recognition}, pages 3156--3164, 2017.

\bibitem{wang2020video}
Heng Wang, Du Tran, Lorenzo Torresani, and Matt Feiszli.
\newblock Video modeling with correlation networks.
\newblock In {\em Proceedings of the IEEE/CVF Conference on Computer Vision and
  Pattern Recognition}, pages 352--361, 2020.

\bibitem{wang2021tdn}
Limin Wang, Zhan Tong, Bin Ji, and Gangshan Wu.
\newblock Tdn: Temporal difference networks for efficient action recognition.
\newblock In {\em Proceedings of the IEEE/CVF Conference on Computer Vision and
  Pattern Recognition}, pages 1895--1904, 2021.

\bibitem{wang2016temporal}
Limin Wang, Yuanjun Xiong, Zhe Wang, Yu Qiao, Dahua Lin, Xiaoou Tang, and Luc
  Van~Gool.
\newblock Temporal segment networks: Towards good practices for deep action
  recognition.
\newblock In {\em European conference on computer vision}, pages 20--36.
  Springer, 2016.

\bibitem{wang2018non}
Xiaolong Wang, Ross Girshick, Abhinav Gupta, and Kaiming He.
\newblock Non-local neural networks.
\newblock In {\em Proceedings of the IEEE conference on computer vision and
  pattern recognition}, pages 7794--7803, 2018.

\bibitem{wang2020symbiotic}
Xiaohan Wang, Yu Wu, Linchao Zhu, and Yi Yang.
\newblock Symbiotic attention with privileged information for egocentric action
  recognition.
\newblock In {\em Proceedings of the AAAI Conference on Artificial
  Intelligence}, volume~34, pages 12249--12256, 2020.

\bibitem{xie2017aggregated}
Saining Xie, Ross Girshick, Piotr Doll{\'a}r, Zhuowen Tu, and Kaiming He.
\newblock Aggregated residual transformations for deep neural networks.
\newblock In {\em Proceedings of the IEEE conference on computer vision and
  pattern recognition}, pages 1492--1500, 2017.

\bibitem{xie2018rethinking}
Saining Xie, Chen Sun, Jonathan Huang, Zhuowen Tu, and Kevin Murphy.
\newblock Rethinking spatiotemporal feature learning: Speed-accuracy trade-offs
  in video classification.
\newblock In {\em Proceedings of the European Conference on Computer Vision
  (ECCV)}, pages 305--321, 2018.

\bibitem{yang2020temporal}
Ceyuan Yang, Yinghao Xu, Jianping Shi, Bo Dai, and Bolei Zhou.
\newblock Temporal pyramid network for action recognition.
\newblock In {\em Proceedings of the IEEE/CVF Conference on Computer Vision and
  Pattern Recognition}, pages 591--600, 2020.

\bibitem{zhang2021token}
Hao Zhang, Yanbin Hao, and Chong-Wah Ngo.
\newblock Token shift transformer for video classification.
\newblock In {\em Proceedings of the 29th ACM International Conference on
  Multimedia}, pages 917--925, 2021.

\bibitem{zhang2018fine}
Hao Zhang and Chong-Wah Ngo.
\newblock A fine granularity object-level representation for event detection
  and recounting.
\newblock {\em IEEE Transactions on Multimedia}, 21(6):1450--1463, 2018.

\bibitem{zhang2020v4d}
Shiwen Zhang, Sheng Guo, Weilin Huang, Matthew~R Scott, and Limin Wang.
\newblock V4d: 4d convolutional neural networks for video-level representation
  learning.
\newblock {\em arXiv preprint arXiv:2002.07442}, 2020.

\bibitem{zhang2018shufflenet}
Xiangyu Zhang, Xinyu Zhou, Mengxiao Lin, and Jian Sun.
\newblock Shufflenet: An extremely efficient convolutional neural network for
  mobile devices.
\newblock In {\em Proceedings of the IEEE conference on computer vision and
  pattern recognition}, pages 6848--6856, 2018.

\bibitem{zhou2018temporal}
Bolei Zhou, Alex Andonian, Aude Oliva, and Antonio Torralba.
\newblock Temporal relational reasoning in videos.
\newblock In {\em Proceedings of the European Conference on Computer Vision
  (ECCV)}, pages 803--818, 2018.

\bibitem{zhu2020cookgan}
Bin Zhu and Chong-Wah Ngo.
\newblock Cookgan: Causality based text-to-image synthesis.
\newblock In {\em Proceedings of the IEEE/CVF Conference on Computer Vision and
  Pattern Recognition}, pages 5519--5527, 2020.

\bibitem{zhu2019r2gan}
Bin Zhu, Chong-Wah Ngo, Jingjing Chen, and Yanbin Hao.
\newblock R2gan: Cross-modal recipe retrieval with generative adversarial
  network.
\newblock In {\em Proceedings of the IEEE/CVF Conference on Computer Vision and
  Pattern Recognition}, pages 11477--11486, 2019.

\end{thebibliography}
}

\clearpage

\newpage
\section*{Appendix}

\begin{figure*}[htbp]
\centering
\includegraphics[width=0.7\textwidth]{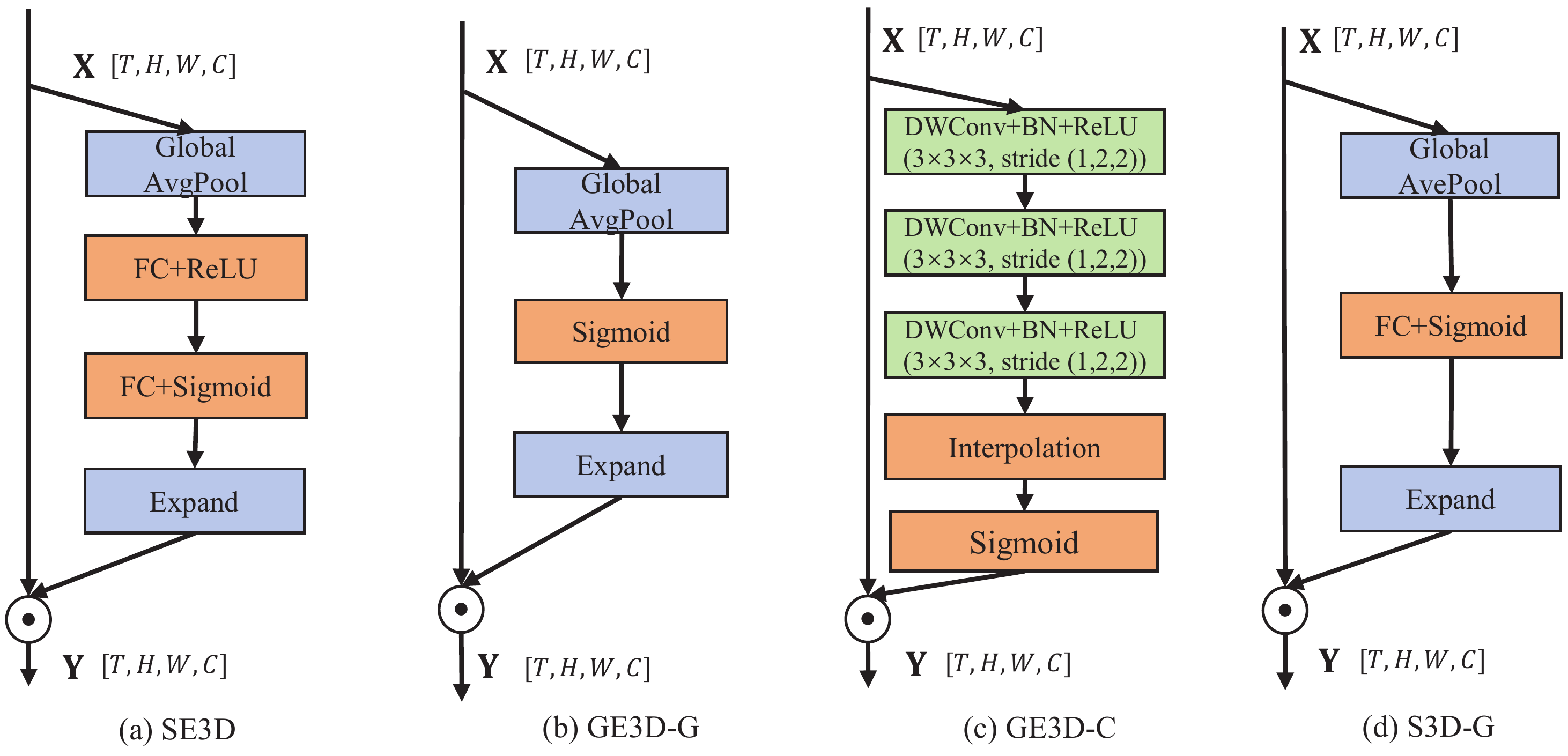}
\caption{Illustration of architectures of SE3D, GE3D-G/C and S3D-G.}
\label{seges3d}
\end{figure*}

This appendix details the architectures of the other calibrators in the ablation study. The results by inserting GCs on different residual stages are provided. Visualization examples of class activation maps on Something-Something V1 dataset are given. Gating weight distributions of different calibrators on Something-Something V1 and Kinetics-400 datasets are shown. More experimental results on datasets EGTEA Gaze+, Diving48 and Basketball-8\&Soccer-10 are presented.

\section*{A. Architectures of SE3D, GE3D-G/C and S3D-G}
Figure \ref{seges3d} shows the detailed architectures of SE3D~\cite{hu2018squeeze}, GE3D-G/C~\cite{hu2018gather} and S3D-G~\cite{xie2018rethinking}. In the implementation, the calibrators are also densely inserted into the TSN and TSM backbones. We report their performances on Something-Something V1 (see ablation study).

\section*{B. Results on different residual stages}
Here, we also investigate which residual stage(s) to add the GC block in ResNet-50 using TSN as backbone. Table \ref{tab:res1to4} compares both a single and multiple GC blocks added to different stages of ResNet. Overall, a GC block significantly improve the performance (19.7\%) of the original TSN. Specifically, the results on the deeper Res3 and Res4 are better than those on Res1 and Res2, and the improvement reaches the highest of 45.9\% on Res3. The possible explanation is that deeper layers can provide more high-level features which are precise for context modeling. But the last residual block Res4 has a small spatial size (7$\times$7) and it limits the precision of spatial information. Moreover, densely incorporating multiple GC blocks into ResNet exhibits better performance than the single version.
\begin{table}[htbp]
		\centering
		\scriptsize
		\begin{tabular}{l|cccccc|c}
			\hline
		    & & \multicolumn{4}{c}{Residual Blocks} &&\\
            \cline{3-6}
			Model & &Res1 &Res2 &Res3 &Res4 &&Top1 (\%)\\
            \midrule[1pt]
            \multirow{7}*{TSN} &&\ding{52} &\ding{55} &\ding{55} &\ding{55} && 42.2 \\
            &&\ding{55} &\ding{52} &\ding{55} &\ding{55} && 43.7 \\
            &&\ding{55} &\ding{55} &\ding{52} &\ding{55} && 45.9 \\
            &&\ding{55} &\ding{55} &\ding{55} &\ding{52} && 44.2 \\
            \cline{2-8}
            &&\ding{52} &\ding{52} &\ding{55} &\ding{55} && 44.3 \\
            &&\ding{52} &\ding{52} &\ding{52} &\ding{55} && 46.7 \\
            &&\ding{52} &\ding{52} &\ding{52} &\ding{52} && \textbf{47.9} \\
            \hline
		\end{tabular}
		\caption{Performance comparison of adding a single or more GC modules to different stages of ResNet-50 on Something-Something V1 dataset.}
		\label{tab:res1to4}
\end{table}

\begin{figure*}[htbp]
\centering
\includegraphics[width=0.95\textwidth]{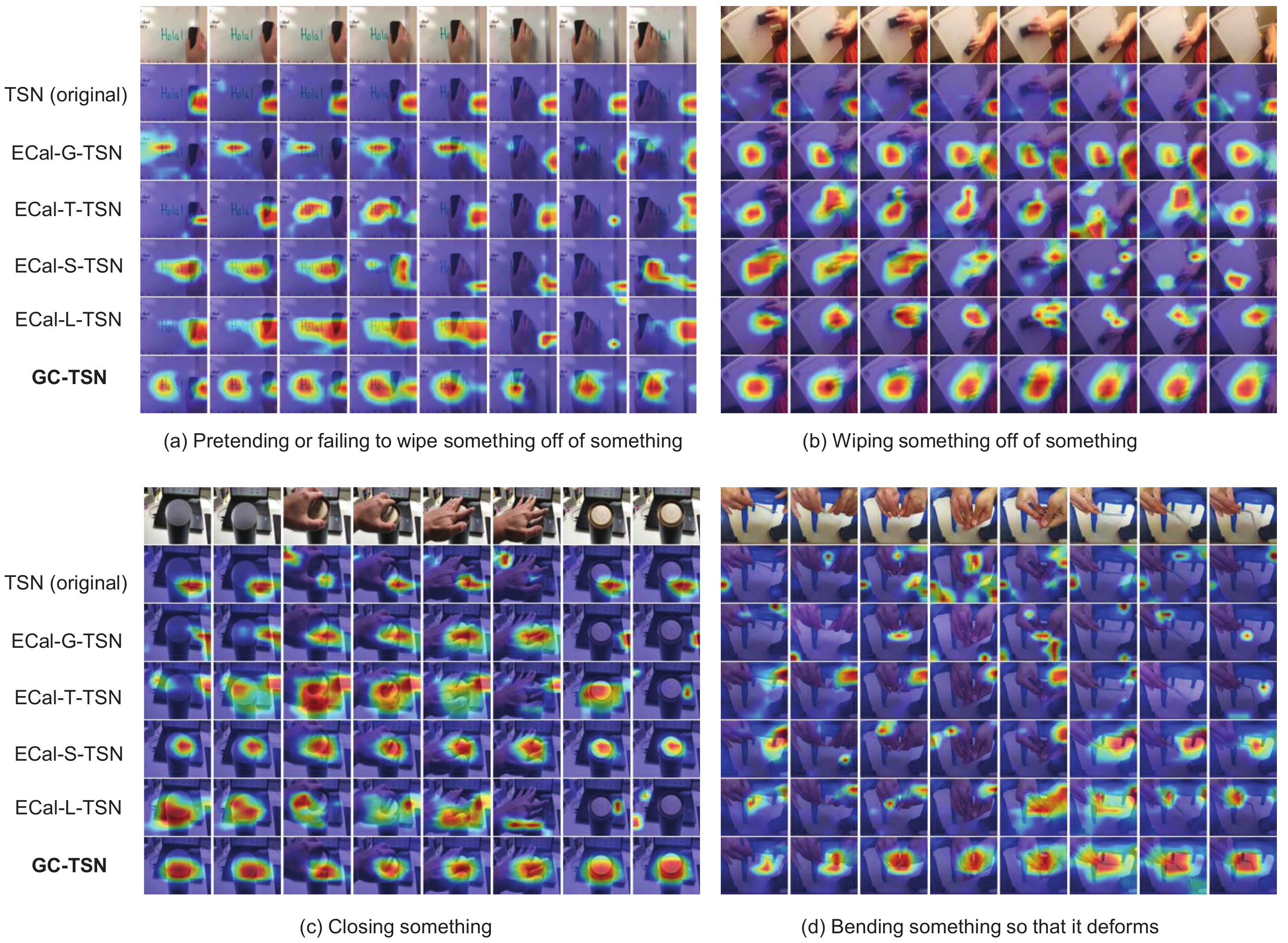}
\caption{Visualization of class activation maps on sample video clips from the Something-Something V1 dataset. The first row presents original frames and each of the other rows presents the visualization results of a model.}
\label{visual}
\end{figure*}

\section*{C. Visualization}
We provide some visualization examples of class activation maps using TSN with ECal-G/T/S/L and GC to clearly show the vital parts they learn. The sampled visualization results are shown in Figure \ref{visual}. In the implementation, we use 8-frame center crops as input and the Grad-CAM \cite{selvaraju2017grad} technique to obtain the heatmaps. These videos are selected from Something-Something V1 dataset. The  categories in Something-Something dataset emphasize not only the short-term interactions between objects (e.g., ``\textit{Wiping something off something}'', ``\textit{Closing something}'' and ``\textit{Bending something}'') but also the long-range dependencies (e.g., ``\textit{Pretending to do}'', ``\textit{Failing to do}'' and ``\textit{Doing something so that it is to be}''). Based on the visualization results, the GC-TSN, which aggregates the four context calibrators ECal-G/T/S/L in parallel, indeed yields more reasonable class activation maps than the original TSN and its variants ECal-TSNs with single-context. 

\section*{D. Gating weight distribution}
\begin{figure}[]
\centering
\includegraphics[width=82mm]{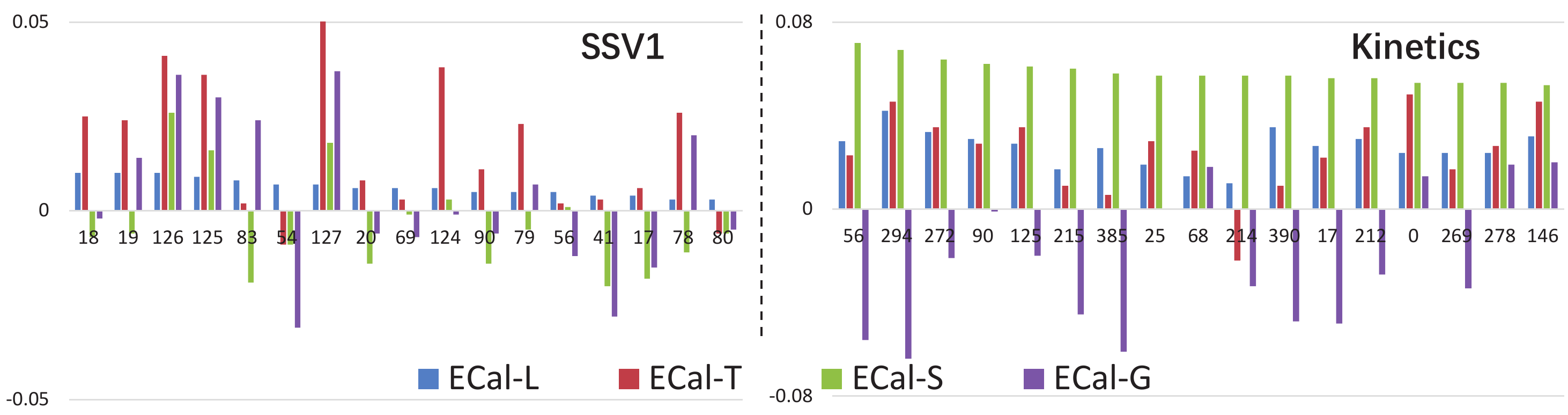}
\vspace{-0.1cm}
\caption{Means of gating weights (before Sigmoid) of ECal-L/T/S/G (GC-TSN) on Something-Something V1 and Kinetics-400.~X-axis: category index.}
\label{gw}
\end{figure}

We calculate the mean of channel weights for each ECal. Fig.\ref{gw} shows the distributions of weights for 17 Something-Something V1 categories involving space-time dynamics and 17 Kinetics-400 categories with less motion variations. The results shows that GC can \textit{inherently} learn the importance of ECals by assigning higher weights for ECal-L/T on Something-Something V1 and ECal-S on Kinetics-400.

\section*{E. Results on other datasets}
In the Appendix, we additionally provide the results on EGTEA Gaze+~\cite{li2018eye}, Diving48~\cite{li2018resound} and  Basketball-8\&Soccer-10~\cite{hao2020compact} datasets. Particularly, the \textbf{EGTEA Gaze+} dataset offers first-person videos, containing 106 non-scripted daily activities occurred in the kitchen. The \textbf{Diving48} dataset consists of 48 unambiguous dive sequence, which requires modeling long-term temporal dynamics. The \textbf{Basketball-8\&Soccer-10} datasets are composed of two datasets for sport classification: Basketball-8 with 8 group activities and Soccer-10 with 10 group activities.


{\bf EGTEA Gaze+}. Table \ref{tab:egtea} shows the results of different methods on the first-vision EGTEA Gaze+ dataset.  Similar observations as on other datasets, GC-Nets contribute significant improvements to their backbones when dealing with the short-term kitchen activities. Specifically, we observe 0.4\%-6.2\% performance increase among the results on the three train/validation splits. This demonstrates that our proposed GC module is generic for short-term temporal modeling. 

\begin{table}[tbp]
\centering
\scriptsize
\begin{tabular}{l|l|c|ccc}
\hline
Method &Backbone & \#Frame &Split1  & Split2 &  Split3 \\ 
\midrule[1pt]
I3D-2stream \cite{li2018eye} & ResNet34 & 24 &55.8  & 53.1 & 53.6  \\
R34-2stream \cite{sudhakaran2018attention} & ResNet34  & 25& 62.2  &61.5 & 58.6 \\ 
SAP~\cite{wang2020symbiotic} & ResNet50  &64 & 64.1  &62.1  &62.0 \\
\hline
TSN (our impl.) & ResNet50 & 8 & 61.6  & 58.5 & 55.2  \\
GST (our impl.) & ResNet50 & 8 & 63.3  & 61.2 & 59.2 \\
TSM (our impl.) & ResNet50 & 8 & 63.5  & 62.8 & 59.5  \\
TDN (our impl.) & ResNet50 & 8 & 63.9  & 60.8 & 60.2  \\
\hline
\textbf{GC-TSN} & ResNet50 & 8 & 66.4  & 64.6 & 61.4 \\
\textbf{GC-GST} & ResNet50 & 8 & 65.5  & 61.6 & 60.6 \\
\textbf{GC-TSM} & ResNet50 & 8 & \textbf{66.5}   & \textbf{66.1} & \textbf{62.6}  \\
\textbf{GC-TDN} & ResNet50 & 8 & 65.0  & 61.8 & 61.0  \\
\hline
\end{tabular}
\caption{Performance (Top-1 accuracy \%) comparison on EGTEA Gaze+ dataset using the official train/validation split 1/2/3.}
\label{tab:egtea}
\end{table}

\begin{table}[]
\centering
\scriptsize
  \begin{tabular}{llcc}
\hline  
\textbf{Method} & \textbf{Backbone} & \textbf{\#Frame} & \textbf{Top-1} \\ 
\midrule[1pt]
SlowFast,16$\times$8 from \cite{bertasius2021space} & ResNet101 &64+16 & 77.6 \\
TimeSformer-HR \cite{bertasius2021space} & Transformer & 16 & 78.0 \\
TimeSformer-L \cite{bertasius2021space} & Transformer & 96 & 81.0 \\
VIMPAC~\cite{tan2021vimpac}  & Transformer & 32 & 85.5 \\
\hline
TSN (our impl.) & ResNet50 & 16 & 79.0 \\
GST (our impl.) & ResNet50 & 16 & 78.9  \\
TSM (our impl.) & ResNet50 & 16 & 83.2  \\
TDN (our impl.) & ResNet50 & 16 & 84.6  \\
\hline
\textbf{GC-TSN}  & ResNet50 & 16 &  86.8  \\
\textbf{GC-GST}  & ResNet50 & 16 &  82.5  \\
\textbf{GC-TSM}  & ResNet50 & 16 & 87.2 \\
\textbf{GC-TDN}  & ResNet50 & 16 & \textbf{87.6} \\
\hline
\end{tabular}
\caption{Performance (Top-1 accuracy \%) comparison on the updated Diving48 dataset using the train/validation split v2.}
\label{tab:diving}
\end{table}

{\bf Diving48}. This dataset is also a ``temporally-hevay'' dataset. Since this newly released dataset version has been thoroughly revised for wrong labels, we re-run the backbones of TSN, TSM, GST and TDN using 16 frames (center crop) as input. Table \ref{tab:diving} shows the performance comparison. Our GC-TDN achieves the highest of 87.6\% Top-1 accuracy, which increases its backbone TDN by 3.0 percentage and is even better than the Transformer-based VIMPAC (85.5\%).

\begin{table}[htbp]
		\centering
		\scriptsize
		\begin{tabular}{l|l|ccrcc}
			\hline
			& & \multicolumn{2}{c}{Basketball-8} & & \multicolumn{2}{c}{Soccer-10}\\
        \cline{3-4}\cline{6-7}
			Model &Pretrain &Validation &Test & &Validation &Test \\
            \midrule[1pt]
            I3D \cite{carreira2017quo} &ImageNet  &--- &75.4 & &--- &88.3 \\
            Nonlocal-I3D \cite{wang2018non} &ImageNet  &--- &77.2 & &--- &88.3 \\
            \hline
            GST \cite{luo2019grouped} (our impl.)  &ImageNet  &78.8 &75.8 & &87.9 &87.6 \\
            \textbf{GC-GST}  &ImageNet  &81.8 &78.4 & &88.3 &88.5 \\
            \hline
            TSN \cite{zhou2018temporal}  &ImageNet  &71.9 &68.5 & &86.2 &83.7 \\
            \textbf{GC-TSN} &ImageNet  &81.8 &78.8 & &89.5 &88.9 \\
            \hline
            TSM \cite{lin2019tsm}   &Kinetics  &77.6 &73.3 & &88.7 &87.9 \\
            +CBA-QSA \cite{hao2020compact} &Kinetics  &---  &78.5  &&---  &89.3 \\
            TSM-NLN \cite{lin2019tsm}  &Kinetics  &---  &76.2  &&---  &88.2 \\
            +CBA-QSA \cite{hao2020compact} &Kinetics &---  &79.5  &&---  &88.7 \\
            \textbf{GC-TSM}  &Kinetics  &\textbf{83.8} &\textbf{80.2} & &\textbf{90.3} &\textbf{89.4} \\
            \hline
            TDN \cite{wang2021tdn}  &ImageNet  &80.3 &78.4 & &86.9 &86.1 \\
            \textbf{GC-TDN}  &ImageNet  &83.0 &79.7 & &87.7 &87.1 \\
            \hline
		\end{tabular}
		\caption{Comparison of performance (Top1 accuracy \%) of differnt methods with 8 frames $\times$ 1 clip input on Sport Highlights datasets. The results of I3D, Nonlocal-I3D and TSM-NLN are cited from \cite{hao2020compact}.}
		\label{tab:res_sport}
\end{table}

{\bf Basketball-8\&Soccer-10}. The fine-grained sport highlights \cite{hao2020compact} take place with various local interactions among offensive players, defensive players and other objects, and the local interactions could be either short-term (e.g., the ``Blocked shot'' highlight in Basketball) or long-term (e.g., the ``Layup'' highlight in Basketball and the ``Shooting and goalkeeping'' highlight in Soccer). This indicates that both the global and local axial contexts can benefit the sport activity recognition. No surprisingly, as shown in Table \ref{tab:res_sport}, all GC-Nets, i.e., GC-TSN, GC-GST, GC-TSM and GC-TDN, consistently boost their base networks, e.g., 68.5\%$\rightarrow$78.8\% for TSN, 75.8\%$\rightarrow$78.4\% for GST, 73.3\%$\rightarrow$80.2\% for TSM and 78.4\%$\rightarrow$79.7\% for GC-TDN. Compared to the similar feature calibration works \cite{wang2018non} and \cite{hao2020compact}, our GC module performs best with the same backbone TSM, obtaining the highest Top1 accuracy 80.2\%/89.4\% on Basketball-8/Soccer-10.

\begin{table*}[htb]
\centering
\begin{tabular}{l|l}
\hline
ID & Name \\ 
\midrule[1pt]
label-1	    &Approaching something with your camera \\ \hline
label-15	&Folding something \\ \hline
label-26	&Lifting a surface with something on it but not enough for it to slide down \\ \hline
label-33	&Moving away from something with your camera \\ \hline
label-37	&Moving something and something away from each other \\ \hline
label-38	&Moving something and something closer to each other \\ \hline
label-44	&Moving something down \\ \hline
label-46	&Moving something up \\ \hline
label-51	&Plugging something into something but pulling it right out as you remove your hand \\ \hline
label-58	&Poking something so that it falls over \\ \hline
label-61	&Pouring something into something until it overflows \\ \hline
label-70	&Pretending to poke something \\ \hline
label-73	&Pretending to put something into something \\ \hline
label-77	&Pretending to put something underneath something \\ \hline
label-78	&Pretending to scoop something up with something \\ \hline
label-81	&Pretending to squeeze something \\ \hline
label-84	&Pretending to throw something \\ \hline
label-88	&Pulling something from right to left \\ \hline
label-91	&Pulling two ends of something but nothing happens \\ \hline
label-106	&Putting something in front of something \\ \hline
label-112	&Putting something onto a slanted surface but it doesn't glide down \\ \hline
label-118	&Putting something that cannot actually stand upright upright on the table, so it falls on its side \\ \hline
label-132	&Something being deflected from something \\ \hline
label-152	&Throwing something \\ \hline
label-156	&Throwing something onto a surface \\ \hline
label-167	&Turning the camera left while filming something \\ \hline
label-168	&Turning the camera right while filming something \\ \hline
label-173	&Unfolding something \\
\hline
\end{tabular}
\caption{Categories of the selected activities in Figure 6.}
\label{tab:lables}
\end{table*}

\section*{F. Video annotations}
Table \ref{tab:lables} lists the selected 28 activity categories from Something-Something V1 dataset used in Figure 6.

\end{document}